\documentclass{article}%
\usepackage{amsmath}
\usepackage{amsfonts}
\usepackage{amssymb}
\usepackage{graphicx}
\usepackage{algorithmic}
\usepackage{algorithm}%
\providecommand{\U}[1]{\protect \rule{.1in}{.1in}}

\begin{document}

\title{SurvBeX: An explanation method of the machine learning survival models based
on the Beran estimator}
\author{Lev V. Utkin and Danila Y. Eremenko and Andrei V. Konstantinov\\Higher School of Artificial Intelligence \\Peter the Great St.Petersburg Polytechnic University\\St.Petersburg, Russia\\e-mail: lev.utkin@gmail.com, danilaeremenko@mail.ru, andrue.konst@gmail.com}
\date{}
\maketitle

\begin{abstract}
An explanation method called SurvBeX is proposed to interpret predictions of
the machine learning survival black-box models. The main idea behind the
method is to use the modified Beran estimator as the surrogate explanation
model. Coefficients, incorporated into Beran estimator, can be regarded as
values of the feature impacts on the black-box model prediction. Following the
well-known LIME method, many points are generated in a local area around an
example of interest. For every generated example, the survival function of the
black-box model is computed, and the survival function of the surrogate model
(the Beran estimator) is constructed as a function of the explanation
coefficients. In order to find the explanation coefficients, it is proposed to
minimize the mean distance between the survival functions of the black-box
model and the Beran estimator produced by the generated examples. Many
numerical experiments with synthetic and real survival data demonstrate the
SurvBeX efficiency and compare the method with the well-known method SurvLIME.
The method is also compared with the method SurvSHAP. The code implementing
SurvBeX is available at: https://github.com/DanilaEremenko/SurvBeX


\end{abstract}

\section{Introduction}

Censored survival data can be met in many applications, including medicine,
the system reliability and safety, economics
\cite{Hosmer-Lemeshow-May-2008,Wang-Li-Reddy-2019}. They consider times to an
event of interest as observations. The event might not have happened during
the period of study \cite{Nezhad-etal-2018}. If the observed survival time is
less than or equal to the true survival time, then the corresponding data is
censored and called right-censoring data. Machine learning models dealing the
survival data are based on applying the framework and concepts of survival
analysis and conditionally divided into parametric, nonparametric and
semiparametric models \cite{Wang-Li-Reddy-2019}.

There are different methods for constituting machine learning survival models.
We consider models which depend on the feature vectors characterizing objects,
whose times to an event are studied, for example, patients, the system
structures, Many survival models are based on using the Cox proportional
hazard method or the Cox model \cite{Cox-1972}, which calculates effects of
observed covariates on the risk of an event occurring under condition that a
linear combination of the instance covariates is assumed in the model. The
success of the Cox model in many applied tasks motivated to generalize the
model in order to consider non-linear functions of covariates or several
regularization constraints for the model parameters
\cite{Haarburger-etal-2018,Katzman-etal-2018,Witten-Tibshirani-2010,Zhu-Yao-Huang-2016}%
. To improve survival analysis and to enhance the prediction accuracy of the
machine learning survival models, other approaches have been developed,
including the SVM approach \cite{Khan-Zubek-2008}, random survival forests
(RSF)
\cite{Ibrahim-etal-2008,Mogensen-etal-2012,Schmid-etal-2016,Wang-Zhou-2017,Wright-etal-2017}%
, survival neural networks
\cite{Haarburger-etal-2018,Katzman-etal-2018,Wiegrebe-etal-23,Zhu-Yao-Huang-2016}%
.

The most survival models can be regarded as black-box models because details
of their functioning are often completely unknown, but their inputs as well as
outcomes may be known for users. As a results, it is difficult to explain how
a prediction corresponding to a certain input is achieved, which features of
the input significantly impact on the prediction. However, the explanation of
predictions should be an important component of intelligent systems especially
in medicine \cite{Holzinger-etal-2019} where a doctor has to understand why
the used machine learning diagnostic model states a diagnosis. At present,
there are many methods which can explain predictions of black-box models
`\cite{Arya-etal-2019,Guidotti-2019,Molnar-2019,Murdoch-etal-19a}. According
to these methods, the explanation means to assign numbers to features, which
quantify the feature impacts on the prediction. The methods can be divided
into local and global explanation methods. Local methods produce explanation
locally around an explained example whereas global methods explain the
black-box model for the whole dataset.

An important local explanation method is the Local Interpretable
Model-agnostic Explanations (LIME) \cite{Ribeiro-etal-2016}, which explains
predictions of a black-box model approximating a model function at a point by
the linear model where coefficients of the linear model can be regarded as the
quantified representation of the feature impacts on the prediction
\cite{Garreau-Luxburg-20a}. According to LIME, the linear model is constructed
by generating randomly synthetic examples around the explained example. It
should be noted that the idea to approximate the black-box model function at a
point by the linear surrogate model is fruitful and can be applied to many
classification and regression tasks.

In most applications, LIME deals with point-valued predictions. However,
predictions of the survival machine learning models are functions, for
example, the survival function (SF), the cumulative hazard function (CHF). In
other words, explanation methods should explain which features have the
significant impact on the form of the SF corresponding to some input object.
To take into account peculiarities of the survival models, a series of the
LIME modifications dealing with survival data and models and called SurvLIME
have been presented in
\cite{Kovalev-Utkin-Kasimov-20a,Kovalev-Utkin-2020c,Utkin-Kovalev-Kasimov-20c}%
. The main idea behind the SurvLIME is to approximate the black-box model at a
point by the Cox model which is based on the linear combination of covariates.
Coefficients of the linear combination can be viewed as measures of the
covariate impact on the prediction in the form of the SF or the CHF.

Many numerical experiments have shown that the proposed SurvLIME models
provide satisfactory results. However, limitations of the Cox model motivated
to search for a more adequate model which could improve the explanation of
survival models. One of the interesting model is the Beran estimator
\cite{Beran-81} which can be regarded as a kernel regression for computing the
SF taking into account the data structure. However, the original Beran
estimator does not contain linear combinations of covariates, which are needed
to develop the linear approximation like the Cox model. Therefore, we proposed
to modify the Beran estimator by incorporating a vector whose elements can be
viewed as the feature impacts on the prediction. The proposed explanation
model using the modified Beran estimator is called SurvBeX (\textbf{Surv}ival
\textbf{Be}ran e\textbf{X}planation). It is similar to SurvLIME, but, in
contrast to SurvLIME, it has several specific elements which distinguish
SurvBeX from SurvLIME. Moreover, SurvBeX is more flexible because it has many
opportunities for modifications. For example, it can be studied with different
kernels which are important elements of the Beran estimator.

Our contributions can be summarized as follows:

\begin{enumerate}
\item A new method for explaining predictions of machine learning survival
models based on applying the modification of the Beran estimator is proposed.

\item An algorithm implementing the method based on solving an unconstrained
optimization problem is proposed. 

\item Various numerical experiments compare the method with the well-known
survival explanation method called SurvLIME under condition of using different
black-box survival models, including the Cox model, the RSF, and the Beran
estimator. SurvBeX is also compared with a very interesting method based on
the survival extension of the method SHAP
\cite{Lundberg-Lee-2017,Strumbel-Kononenko-2010}, which is called SurvSHAP(t)
\cite{Krzyzinski-etal-23}. The corresponding code implementing the proposed
method is publicly available at: https://github.com/DanilaEremenko/SurvBeX.
\end{enumerate}

Numerical results show that SurvBeX provides outperforming results in
comparison with SurvLIME and SurvSHAP(t).

The paper is organized as follows. Related work can be found in Section 2. A
short description of basic concepts of survival analysis and the explanation
methods, including the Cox model, the original Beran estimator, LIME,
SurvLIME, is given in Section 3. A general idea of SurvBeX is provided in
Section 4. The formal derivation of the optimization problem for implementing
SurvBeX and its simplification can be found in Section 5. Numerical
experiments with synthetic data are provided in Section 6. Similar numerical
experiments with real data are given in Section 7. Concluding remarks are
provided in Section 8.

\section{Related work}

\textbf{Local explanation methods.} Due to the importance of explaining and
interpreting the machine learning model predictions, many local explanation
methods have been developed and studied. First of all, we have to point out
the original LIME \cite{Ribeiro-etal-2016} and its various modifications,
including ALIME \cite{Shankaranarayana-Runje-2019}, NormLIME
\cite{Ahern-etal-2019}, DLIME \cite{Zafar-Khan-2019}, Anchor LIME
\cite{Ribeiro-etal-2018}, LIME-SUP \cite{Hu-Chen-Nair-Sudjianto-2018},
LIME-Aleph \cite{Rabold-etal-2019}, GraphLIME \cite{Huang-Yamada-etal-2020},
Rank-LIME \cite{Chowdhury-etal-22}, s-LIME \cite{Gaudel-etal-22}.

Along with LIME, another method of local explanation, called SHAP
\cite{Strumbel-Kononenko-2010}, is also actively used. It is based on applying
a game-theoretic approach for optimizing a regression loss function using
Shapley values \cite{Lundberg-Lee-2017}. Due to the strong theoretical
justification of SHAP, a lot of its modifications also have been proposed, for
example, SHAP-E \cite{Jun-Nichol-23}, SHAP-IQ \cite{Fumagalli-etal-23}, K-SHAP
\cite{Coletta-etal-23}, MM-SHAP \cite{Parcalabescu-Frank-22}, Latent SHAP
\cite{Bitto-etal-22}, X-SHAP \cite{Bouneder-etal-20}, Random SHAP
\cite{Utkin-Konstantinov-22n}.

Another interesting class of explanation methods is based on using ideas
behind the Generalized Additive Model (GAM) \cite{Hastie-Tibshirani-1990}.
Following this mode, the explainable boosting machine was proposed in
\cite{Nori-etal-19} and \cite{Chang-Tan-etal-2020}. Another model based on GAM
is the Neural Additive Model (NAM) was proposed in \cite{Agarwal-etal-20}. NAM
can be viewed as a neural network implementation of GAM where the network
consists of a linear combination of \textquotedblleft small\textquotedblright%
\ neural subnetworks having single inputs corresponding to each feature, i.e.
a single feature is fed to each subnetwork producing a shape function of a
feature. Similar approaches using neural networks for implementing GAM and
performing shape functions called GAMI-Net and the Adaptive Explainable Neural
Networks (AxNNs) were proposed in \cite{Yang-Zhang-Sudjianto-20} and
\cite{Chen-Vaughan-etal-20}, respectively,

Critical and comprehensive surveys devoted to the local explanation methods
can be found in
\cite{Adadi-Berrada-2018,Arrieta-etal-2020,Bodria-etal-23,Burkart-Huber-21,Carvalho-etal-2019,Cwiling-etal-23,Guidotti-2019,Molnar-2019,Murdoch-etal-19a,Rudin-2019}%
.

\textbf{Machine learning models in survival analysis}. Many machine learning
models have the survival extension, i.e. they are modified in order to deal
with survival data or with censored data. A detailed review of survival models
is presented in \cite{Wang-Li-Reddy-2019}. The survival machine learning
models can be divided into three groups. The first group consists of models
which are based on the well-known statistical models, including the Cox model,
The Kaplan-Meier models, etc. Several models from the first group use the
Nadaraya-Watson regression for estimating SFs and other concepts of survival
analysis
\cite{Bobrowski-etal-15,Gneyou-14,Pelaez-Cao-Vilar-22,Selingerova-etal-21,Tutz-Pritscher-96}%
. Kernel-based estimators of SFs in survival analysis are studied in
\cite{Pelaez-Cao-Vilar-21}. The second group consists of models which mainly
adapt the original machine learning models to survival data. These are the SVM
approach \cite{Widodo-Yang-2011}, the random survival forests
\cite{Ibrahim-etal-2008,Wang-Zhou-2017,Wright-etal-2017}, survival neural
networks \cite{Katzman-etal-2018,Wiegrebe-etal-23}. The third group is some
combinations of the models from the first and the second groups. Examples of
these models are modifications of the Cox model relaxing the linear
relationship assumption accepted in the Cox model, including neural networks
instead of the linear relationship
\cite{Faraggi-Simon-1995,Haarburger-etal-2018}, the Lasso modifications
\cite{Kaneko-etal-2015,Ternes-etal-2016,Witten-Tibshirani-2010}.

Most aforementioned survival models can be viewed as black-box models whose
predictions should be explained in many applications. This motivates to
develop the explanation methods which deal with survival data.

\textbf{Explanation methods in survival analysis. }One of the problems
encountered the local explanation in survival analysis is to explain the
function-valued predictions instead of the point-valued predictions which are
explained by many methods. In survival analysis, predictions are usually SFs
or CHFs which are functions of time. A new approach to deal with
function-valued predictions is SurvLIME
\cite{Kovalev-Utkin-Kasimov-20a,Kovalev-Utkin-2020c,Utkin-Kovalev-Kasimov-20c}%
. Its modification and the corresponding software is called SurvLIMEpy
\cite{Pachon-Garcia-etal-23}. An extension to SurvLIME called SurvNAM has been
proposed in \cite{Utkin-Satyukov-Konstantinov-22}. SurvNAM\ can be also viewed
as an extension to NAM \cite{Agarwal-etal-20}. An extension of the NAM for
survival analysis with EHR Data was presented in \cite{Peroni-etal-22}.

A novel, easily deployable approach, called EXplainable CEnsored Learning, to
iteratively exploit critical variables in the framework of survival analysis
was developed in \cite{Wu-Peng-etal-22}. The authors of this work aim to find
critical features with long-term prognostic values. The explainable machine
learning, which provides insights in breast cancer survival, is considered in
\cite{Moncada-Torres-etal-21}. An interesting extension of the SHAP to its
application in survival analysis, called SurvSHAP(t) has been proposed in
\cite{Krzyzinski-etal-23}. The authors of the work introduce the first
time-dependent explanation for interpreting machine learning survival models.
The important feature of the SurvSHAP(t) method is that it explains the
predicted SF.

\section{Elements of survival analysis}

\subsection{Basic concepts}

An example (object) in survival analysis is usually represented by a triplet
$(\mathbf{x}_{i},\delta_{i},T_{i})$, where $\mathbf{x}_{i}^{\mathrm{T}%
}=(x_{i1},...,x_{id})$ is the vector of the example features; $T_{i}$ is time
to event of interest for the $i$-th example. If the event of interest is
observed, then $T_{i}$ is the time between a baseline time and the time of
event happening. In this case, an uncensored observation takes place and
$\delta_{i}=1$. Another case is when the event of interest is not observed.
Then $T_{i}$ is the time between the baseline time and the end of the
observation. In this case, a censored observation takes place and $\delta
_{i}=0$. There are different types of censored observations. We will consider
only right-censoring, where the observed survival time is less than or equal
to the true survival time \cite{Hosmer-Lemeshow-May-2008}. Given a training
set $\mathcal{A}$ consisting of $n$ triplets $(\mathbf{x}_{i},\delta_{i}%
,T_{i})$, $i=1,...,n$, the goal of survival analysis is to estimate the time
to the event of interest $T$ for a new example $\mathbf{x}$ by using
$\mathcal{A}$.

Key concepts in survival analysis are SFs, hazard functions, and the CHF,
which describe probability distributions of the event times. The SF denoted as
$S(t|\mathbf{x})$ is the probability of surviving up to time $t$, that is
$S(t|\mathbf{x})=\Pr \{T>t|\mathbf{x}\}$. The hazard function $h(t|\mathbf{x})$
is the rate of the event at time $t$ given that no event occurred before time
$t$. The hazard function can be expressed through the SF as follows
\cite{Hosmer-Lemeshow-May-2008}:
\begin{equation}
h(t|\mathbf{x})=-\frac{\mathrm{d}}{\mathrm{d}t}\ln S(t|\mathbf{x}).
\end{equation}

The CHF denoted as $H(t|\mathbf{x})$ is defined as the integral of the hazard
function $h(t|\mathbf{x})$ and can be interpreted as the probability of the
event at time $t$ given survival until time $t$, that is
\begin{equation}
H(t|\mathbf{x})=\int_{-\infty}^{t}h(z|\mathbf{x})dz.
\end{equation}

An important relationship between the SF and the CHF is
\begin{equation}
S(t|\mathbf{x})=\exp \left(  -H(t|\mathbf{x})\right)  .
\end{equation}

To compare survival models, the C-index proposed by Harrell et al.
\cite{Harrell-etal-1982} is used. It estimates how good a survival model is at
ranking survival times. It estimates the probability that, in a randomly
selected pair of examples, the example that fails first had a worst predicted
outcome. In fact, this is the probability that the event times of a pair of
examples are correctly ranking.

C-index has different forms. One of the forms is \cite{Uno-etal-11}:
\[
C=\frac{\sum \nolimits_{i,j}\mathbf{1}[T_{i}<T_{j}]\cdot \mathbf{1}[\widehat
{T}_{i}<\widehat{T}_{j}]\cdot \delta_{i}}{\sum \nolimits_{i,j}\mathbf{1}%
[T_{i}<T_{j}]\cdot \delta_{i}},
\]
where $\widehat{T}_{i}$ and $\widehat{T}_{j}$ are the predicted survival durations.

\subsection{The Cox model}

The Cox proportional hazards model is used as an important component in
SurvLIME. Therefore, we provide some details of the model. According to the
Cox model, the hazard function at time $t$ given vector $\mathbf{x}$ is
defined as \cite{Cox-1972,Hosmer-Lemeshow-May-2008}:
\begin{equation}
h(t|\mathbf{x},\mathbf{b})=h_{0}(t)\exp \left(  \mathbf{b}^{\mathrm{T}%
}\mathbf{x}\right)  . \label{SurvLIME1_10}%
\end{equation}

Here $h_{0}(t)$ is a baseline hazard function which does not depend on the
vector $\mathbf{x}$ and the vector $\mathbf{b}$; $\mathbf{b}^{\mathrm{T}%
}=(b_{1},...,b_{m})$ is an unknown vector of the regression coefficients or
the model parameters. The baseline hazard function represents the hazard when
all of the covariates are equal to zero.

The SF in the framework of the Cox model is
\begin{equation}
S(t|\mathbf{x},\mathbf{b})=\exp(-H_{0}(t)\exp \left(  \mathbf{b}^{\mathrm{T}%
}\mathbf{x}\right)  )=\left(  S_{0}(t)\right)  ^{\exp \left(  \mathbf{b}%
^{\mathrm{T}}\mathbf{x}\right)  }. \label{Cox_SF}%
\end{equation}

Here $H_{0}(t)$ is the baseline CHF; $S_{0}(t)$ is the baseline SF which is
defined as the survival function for zero feature vector $\mathbf{x}%
^{\mathrm{T}}=(0,...,0)$. The baseline SF and CHF do not depend on
$\mathbf{x}$ and $\mathbf{b}$.

One of the ways to compute the parameters $\mathbf{b}$ is to use the partial
likelihood defined as follows \cite{Cox-1972}:
\begin{equation}
L(\mathbf{b})=\prod_{j=1}^{n}\left[  \frac{\exp(\mathbf{b}^{\mathrm{T}%
}\mathbf{x}_{j})}{\sum_{i\in R_{j}}\exp(\mathbf{b}^{\mathrm{T}}\mathbf{x}%
_{i})}\right]  ^{\delta_{j}},
\end{equation}
where $R_{j}$ is the set of objects which are at risk at time $t_{j}$. The
term \textquotedblleft at risk at time $t$\textquotedblright \ means objects
for which the event of interest happens at time $t$ or later. Only uncensored
observations are used in the likelihood function.

\subsection{The Beran estimator}

Given the dataset $\mathcal{A}$, the SF can be estimated by using the Beran
estimator \cite{Beran-81} as follows:
\begin{equation}
S_{B}(t|\mathbf{x},\mathcal{A})=\prod_{t_{i}\leq t}\left \{  1-\frac
{\alpha(\mathbf{x},\mathbf{x}_{i})}{1-\sum_{j=1}^{i-1}\alpha(\mathbf{x}%
,\mathbf{x}_{j})}\right \}  ^{\delta_{i}},\label{Beran_est}%
\end{equation}
where time moments are ordered; the weight $\alpha(\mathbf{x},\mathbf{x}_{i})$
conforms with relevance of the $i$-th example $\mathbf{x}_{i}$ to the vector
$\mathbf{x}$ and can be defined through kernels as%
\begin{equation}
\alpha(\mathbf{x},\mathbf{x}_{i})=\frac{K(\mathbf{x},\mathbf{x}_{i})}%
{\sum_{j=1}^{n}K(\mathbf{x},\mathbf{x}_{j})}.
\end{equation}

If we use the Gaussian kernel, then the weights $\alpha(\mathbf{x}%
,\mathbf{x}_{i})$ are of the form:
\begin{equation}
\alpha(\mathbf{x},\mathbf{x}_{i})=\text{\textrm{softmax}}\left(
-\frac{\left \Vert \mathbf{x}-\mathbf{x}_{i}\right \Vert ^{2}}{\tau}\right)  ,
\end{equation}
where $\tau$ is a tuning (temperature) parameter.

Below another representation of the Beran estimator will be used:%
\begin{equation}
S_{B}(t|\mathbf{x},\mathcal{A})=\prod_{t_{i}\leq t}\left \{  \frac{1-\sum
_{j=1}^{i}\alpha(\mathbf{x},\mathbf{x}_{j})}{1-\sum_{j=1}^{i-1}\alpha
(\mathbf{x},\mathbf{x}_{j})}\right \}  ^{\delta_{i}}.
\end{equation}

The Beran estimator can be regarded as a generalization of the Kaplan-Meier
estimator \cite{Wang-Li-Reddy-2019} because it is reduced to the Kaplan-Meier
estimator if the weights $\alpha(\mathbf{x},\mathbf{x}_{i})$ take values
$\alpha(\mathbf{x},\mathbf{x}_{i})=1/n$ for all $i=1,...,n$. The product in
(\ref{Beran_est}) takes into account only uncensored observations whereas the
weights are normalized by using uncensored as well as censored observations.

\subsection{LIME}

LIME aims to approximate a black-box model implementing a function
$q(\mathbf{x)}$ by a linear function $g(\mathbf{x)}$ in the vicinity of the
explained point $\mathbf{x}$ \cite{Ribeiro-etal-2016}. Generally, the function
$g(\mathbf{x)}$ may be arbitrary from a set $G$ of explainable functions. LIME
proposes to perturb new examples and obtain the corresponding predictions for
the examples by means of the black-box model. This set of the examples jointly
with the predictions forms a dataset for learning the function $g(\mathbf{x)}%
$. To take into account different distances from $\mathbf{x}$ and the
generated examples, weights $w_{\mathbf{x}}$ are assigned in accordance with
the distances by using a kernel function.

An explanation (local surrogate) model is trained by solving the following
optimization problem:
\begin{equation}
\arg \min_{g\in G}L(q,g,w_{\mathbf{x}})+\Phi(g).
\end{equation}

Here $L$ is a loss function, for example, mean squared error (MSE), which
measures how the explanation is close to the prediction of the black-box
model; $\Phi(g)$ is the model complexity.

If $g(\mathbf{x)}$ is linear, then the corresponding prediction is explained
by analyzing values of coefficients of $g(\mathbf{x)}$.

\subsection{SurvLIME}

SurvLIME is a modification of LIME when the black-box model is survival
\cite{Kovalev-Utkin-Kasimov-20a}. In contrast to the original LIME, SurvLIME
approximates the black-box model with the Cox model.

Suppose there is a training set $\mathcal{A}$ and a black-box model. The model
prediction is represented in the form of the CHF $H(t|\mathbf{x})$ for every
new example $\mathbf{x}$. SurvLIME approximates the black-box model prediction
by the Cox model prediction with the CHF $H_{\text{Cox}}(t|\mathbf{x}%
,\mathbf{b})$. Values of parameters $\mathbf{b}$ can be regarded as
quantitative impacts on the prediction $H(t|\mathbf{x})$. They are unknown and
have to be found by means of the approximation of models. Optimal coefficients
$\mathbf{b}$ make the distance between CHFs $H(t|\mathbf{x})$ and
$H_{\text{Cox}}(t|\mathbf{x},\mathbf{b})$ for the example $\mathbf{x}$ as
small as possible.

In order to implement the approximation at point $\mathbf{x}$, many nearest
examples $\mathbf{x}_{k}$ are generated in a local area around $\mathbf{x}$.
The CHF $H(t|\mathbf{x}_{k})$ predicted by the black-box model is computed for
every generated $\mathbf{x}_{k}$. Optimal values of $\mathbf{b}$ can be found
by minimizing the weighted average distance between every pair of CHFs
$H(t|\mathbf{x}_{k})$ and $H_{\text{Cox}}(t|\mathbf{x}_{k},\mathbf{b})$ over
all generated points $\mathbf{x}_{k}$. Weight $w_{k}$ depends on the distance
between points $\mathbf{x}_{k}$ and $\mathbf{x}$. Smaller distances between
$\mathbf{x}_{k}$ and $\mathbf{x}$ define larger weights of distances between
CHFs. The distance metric between CHFs defines the corresponding optimization
problem for computing optimal coefficients $\mathbf{b}$. Therefore, SurvLIME
\cite{Kovalev-Utkin-Kasimov-20a} uses the $l_{2}$-norm applied to logarithms
of CHFs $H(t|\mathbf{x}_{k})$ and $H_{\text{Cox}}(t|\mathbf{x}_{k}%
,\mathbf{b})$, which leads to a simple convex optimization problem with
variables $\mathbf{b}$.

One of the limitations of SurvLIME is caused by the underlying Cox model: even
in small neighborhoods in the feature space, the proportional hazards
assumption may be violated, leading to an inaccurate surrogate model. Another
limitation is the linear relationship of covariates in the Cox model, which
may be a reason for the inadequate approximation of black-box models by
SurvLIME. In order to overcome the aforementioned SurvLIME drawbacks, it is
proposed to replace the Cox model by the Beran estimator for explaining
predictions produced by a black-box. In other words, it is proposed to
approximate the prediction of the black-box model by means of the Beran estimator.

\section{A general idea of SurvBeX}

Given a training set $\mathcal{A}$ and a black-box model which is explained,
for every new example $\mathbf{x}$, the black-box model produces the
corresponding output in the form of the SF $S(t|\mathbf{x})$. A general
algorithm of SurvBeX is similar to SurvLIME. Suppose we have a new example
$\mathbf{x}$ and the black-box produces a prediction in the form of SF
$S(t|\mathbf{x})$. Let us incorporate parameters $\mathbf{b}=(b_{1}%
,...,b_{d})$ into the Beran estimator (\ref{Beran_est}) as follows:%
\begin{equation}
\alpha(\mathbf{x},\mathbf{x}_{i},\mathbf{b})=\text{\textrm{softmax}}\left(
-\frac{\left \Vert \mathbf{b}\odot \left(  \mathbf{x}-\mathbf{x}_{i}\right)
\right \Vert ^{2}}{\tau}\right)  .\label{Beran_est_32}%
\end{equation}

Here $\mathbf{b}\odot \mathbf{x}$ is the dot product of two vectors. It can be
seen from the above that each element of $\mathbf{b}$ can be regarded as the
quantitative impact of the corresponding feature on the prediction
$S_{B}(t|\mathbf{x})$ which will be denoted as $S_{B}(t|\mathbf{x}%
,\mathbf{b})$ to indicate the dependence of the SF on parameters $\mathbf{b}$.
If we approximate the function $S(t|\mathbf{x})$ by the function
$S_{B}(t|\mathbf{x},\mathbf{b})$, then coefficients $\mathbf{b}$ show which
features in $\mathbf{x}$ are the most important. Actually, elements of
$\mathbf{b}$ weigh the difference between features $x_{j}$ and $x_{ij}$ of
vectors $\mathbf{x}$ and $\mathbf{x}_{i}$, respectively. However, it follows
from the Beran estimator that the predicted SF $S_{B}(t|\mathbf{x})$ depends
on differences of vectors $\mathbf{x}$ and $\mathbf{x}_{i}$, but not on the
vector $\mathbf{x}$ itself. Therefore, we use the above representation for
$\alpha(\mathbf{x},\mathbf{x}_{i},\mathbf{b})$ in (\ref{Beran_est}).

The remaining part of the algorithm does not substantially differ from
SurvLIME. Suppose that the black-box model has been trained on the dataset
$\mathcal{A}$. In order to find vector $\mathbf{b}$ in accordance with LIME or
SurvLIME, we generate $N$ points $\mathbf{z}_{1},...,\mathbf{z}_{N}$ in a
local area around $\mathbf{x}$. Different notations for training and generated
examples are used in order to distinguish the corresponding sets of data. In
contrast to vectors $\mathbf{x}_{1},...,\mathbf{x}_{n}$ from $\mathcal{A}$,
which train the black-box model and the Beran estimator, points $\mathbf{z}%
_{k}$ can be viewed as tested examples for the black-box model and for the
Beran estimator. For every point $\mathbf{z}_{k}$, the SF $S(t|\mathbf{z}%
_{k})$ is determined as a prediction of the black-box model. The SF
$S_{B}(t|\mathbf{z}_{k},\mathbf{b})$ is represented by using (\ref{Beran_est})
as a function of the parameter  vector $\mathbf{b}$ which is unknown at this
stage. So, we obtain $N$ SFs $S(t|\mathbf{z}_{k})$ and the same number of SFs
$S_{B}(t|\mathbf{z}_{k},\mathbf{b})$. It is important to note that the Beran
estimator uses the training dataset $\mathcal{A}$ to make predictions for the
generated points $\mathbf{z}_{k}$. 

In order to approximate functions $S(t|\mathbf{z}_{k})$ and $S_{B}%
(t|\mathbf{z}_{k},\mathbf{b})$ and to compute the vector $\mathbf{b}$, we
minimize the average distance between the functions over the vector
$\mathbf{b}$. The solution to the optimization problem is the optimal vector
$\mathbf{b}^{opt}$ which explains the prediction $S(t|\mathbf{x})$
corresponding to the example $\mathbf{x}$. At that, we additionaly assign
weights $w_{1},...,w_{N}$ to examples $\mathbf{z}_{1},...,\mathbf{z}_{N}$ such
that the weight $w_{k}$ is defined by the distance between vectors
$\mathbf{z}_{k}$ and $\mathbf{x}$. Smaller distances between $\mathbf{z}_{k}$
and $\mathbf{x}$ produce larger weights of distances between SFs
$S(t|\mathbf{z}_{k})$ and $S_{B}(t|\mathbf{z}_{k},\mathbf{b})$. Weights can be
calculated by using the standard normalized kernel $K^{\ast}(\mathbf{z}%
_{k},\mathbf{x})$. It should distinguish kernels $K^{\ast}$ and $K$. The
kernel $K^{\ast}$ is for computing weights of generated examples, but the
kernel $K$ is for computing weights $\alpha(\mathbf{x},\mathbf{x}%
_{i},\mathbf{b})$ from the Beran estimator. Weights $w_{k}$ are defined in the
same way as in LIME.%

\begin{figure}
[ptb]
\begin{center}
\includegraphics[
height=4.0482in,
width=5.217in
]%
{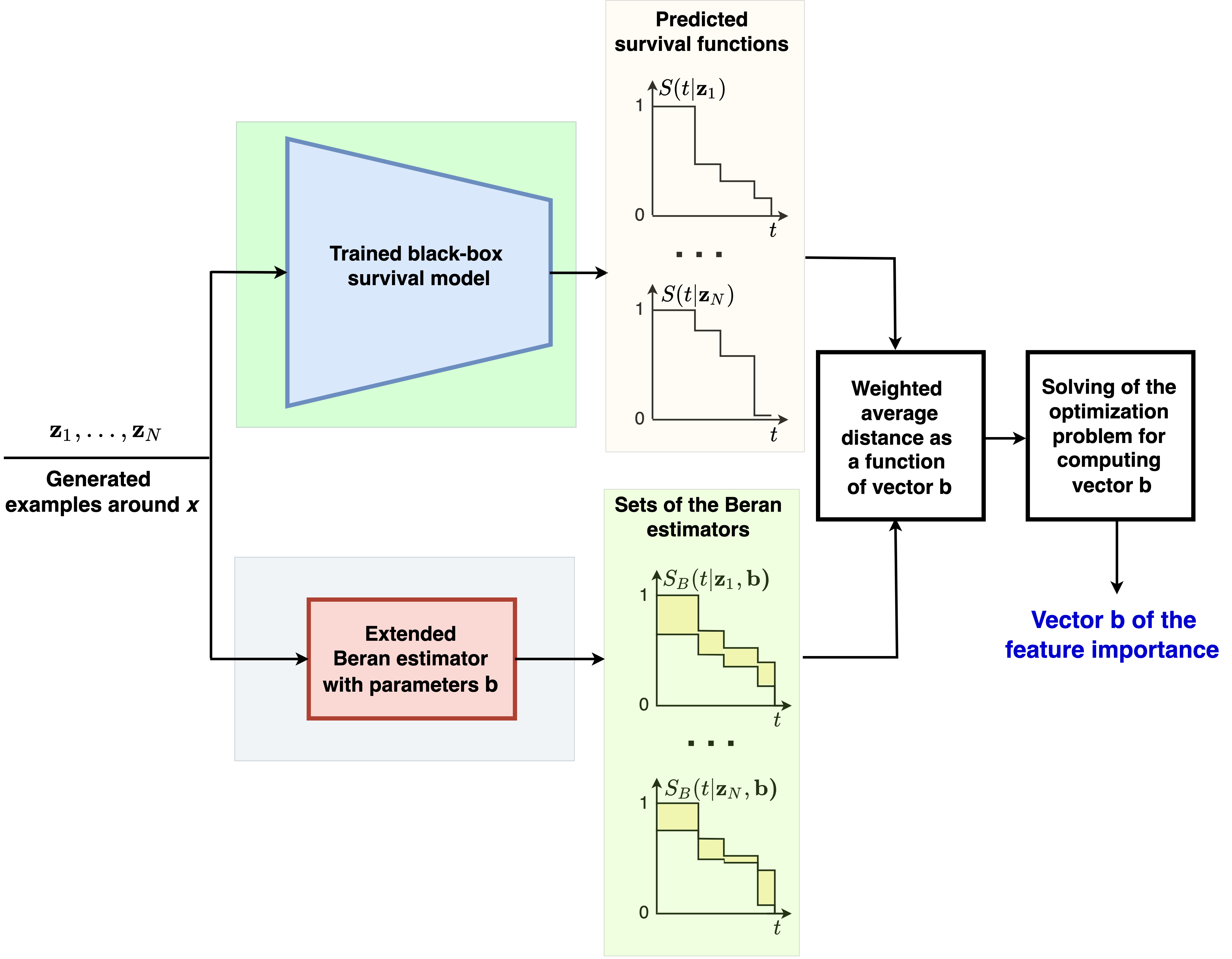}%
\caption{A general scheme of SurvBeX}%
\label{f:beran_explain_1}%
\end{center}
\end{figure}

A general scheme of SurvBeX is depicted in Fig. \ref{f:beran_explain_1}. It
can be seen from Fig. \ref{f:beran_explain_1} that two models (the trained
black-box survival model and the extended Beran estimator with parameters
$\mathbf{b}$) produce SFs corresponding to the generated points $\mathbf{z}%
_{1},...,\mathbf{z}_{N}$. Moreover, SFs produced by the Beran estimator can be
regarded as a family of functions parametrized by $\mathbf{b}$. They are
obtained in the \textit{implicit} form as functions of $\mathbf{b}$. The
objective function for computing $\mathbf{b}^{opt}$ is the weighted average
distance between SFs $S(t|\mathbf{z}_{k})$ and $S_{B}(t|\mathbf{z}%
_{k},\mathbf{b})$.

Only an idea of SurvBeX is presented above. A formal description of the method
which contains details of its implementation is provided in the next section.

\section{Formal optimization problem statement}

According to the general description of SurvBeX, $N$ examples are randomly
generated with weights $w_{j}$ to construct an approximation of the predicted
SF by the SF estimated by means of the Beran estimator at point $\mathbf{x}$.
If we have the set $\mathcal{A}$ of $n$ training examples $(\mathbf{x}%
_{i},\delta_{i},T_{i})$ and the set of $N$ generated feature vectors
$\mathbf{z}_{j}$ in a local area around the explained example $\mathbf{x}$,
then the objective function of the optimization problem for computing optimal
vector $\mathbf{b}$ is the weighted average distance between $N$ pairs of SFs
$S(t|\mathbf{z}_{j})$ and $S_{B}(t|\mathbf{z}_{j},\mathbf{b})$, with weights
$w_{j}$, $j=1,...,N$, which can be written as follows:
\begin{equation}
\min_{\mathbf{b}}\frac{1}{N}\sum_{j=1}^{N}w_{j}\cdot D\left(  S(t|\mathbf{z}%
_{j}),S_{B}(t|\mathbf{z}_{j},\mathbf{b})\right)  .
\end{equation}

Here $D\left(  \cdot,\cdot \right)  $ is a distance between two functions.
Since many distance metrics are based on the norms $l_{s}$, $s=1,...,\infty$,
then we can rewrite the optimization problem for these distance metrics
\begin{equation}
\min_{\mathbf{b}}\frac{1}{N}\sum_{j=1}^{N}w_{j}\left \Vert S(t|\mathbf{z}%
_{j})-S_{B}(t|\mathbf{z}_{j},\mathbf{b})\right \Vert _{s}^{s},
\end{equation}
which in the case of finite $s$ is equivalent to
\begin{equation}
\min_{\mathbf{b}}\sum_{j=1}^{N}w_{j}\int_{0}^{\infty}\left \vert S(t|\mathbf{z}%
_{j})-S_{B}(t|\mathbf{z}_{j},\mathbf{b})\right \vert ^{s}.
\end{equation}

Let $t_{0}<t_{1}<...<t_{n}$ be ordered times to an event of interest from the
set $\{T_{1},...,T_{n}\}$, where $t_{0}=0$ and $t_{n}=\max_{k=1,...,n}T_{k}$.
It is assumed that all times are different. The times $t_{0},t_{1},...,t_{n}$
split the interval $\Omega=[t_{0},t_{n}]$ into $n$ non-intersecting
subintervals $\Omega_{1},...,\Omega_{n}$ such that $\Omega_{k}=[t_{k-1}%
,t_{k})$, $k=1,...,n$.

The black-box model maps the feature vectors $\mathbf{x}\in \mathbb{R}^{d}$
into piecewise constant SF $S(t|\mathbf{x})$ which can be represented as the
sum
\begin{equation}
S(t|\mathbf{x})=\sum_{k=1}^{n}S^{(k)}(\mathbf{x})\cdot \chi_{\Omega_{k}}(t),
\end{equation}
where $\chi_{\Omega_{k}}(t)$ is the indicator function taking value $1$ if
$t\in \Omega_{k}$, and value $0$ if $t\notin \Omega_{k}$; $S^{(k)}(\mathbf{x})$
is the SF for $t\in \Omega_{k}$.

The same representation can be written for the SF $S_{B}(t|\mathbf{x}%
,\mathbf{b})$ as well as for the difference of the SFs. We also assume that
$S(t|\mathbf{x})=0$ for $t>t_{n}$. Then the objective function in the
optimization problem for computing $\mathbf{b}$ can be formulated as
\begin{align}
&  \sum_{j=1}^{N}w_{j}\int_{\Omega}\left \vert S(t|\mathbf{z}_{j}%
)-S_{B}(t|\mathbf{z}_{j},\mathbf{b})\right \vert ^{s}\nonumber \\
&  =\sum_{j=1}^{N}w_{j}\sum_{i=1}^{n}\left \vert S^{(i)}(\mathbf{z}_{j}%
)-S_{B}^{(i)}(\mathbf{z}_{j},\mathbf{b})\right \vert ^{s}\left(  t_{i}%
-t_{i-1}\right)  .\label{Beran_est_16}%
\end{align}

Similarly, we can state a problem to minimize the difference between
logarithms of SFs in order to simplify the optimization problem. Let us
introduce the condition $S^{(i)}(\mathbf{x})\geq \varepsilon>0$ for all
$i=1,...,n$. Then the optimizatin problem can also be formulated through
logarithms:
\begin{equation}
\mathbf{b}=\arg \min_{\mathbf{b}}\sum_{j=1}^{N}w_{j}\sum_{i=1}^{n}\left \vert
\ln S^{(i)}(\mathbf{z}_{j})-\ln S_{B}^{(i)}(\mathbf{z}_{j},\mathbf{b}%
)\right \vert ^{s}\left(  t_{i}-t_{i-1}\right)  ,\label{Beran_est_18}%
\end{equation}
where
\begin{equation}
\ln S_{B}^{(i)}(\mathbf{z}_{j},\mathbf{b})=\sum_{t_{l}\leq t_{i}}\delta
_{l}\left[  \ln \left(  1-\sum_{k=1}^{l}\alpha(\mathbf{z}_{j},\mathbf{x}%
_{k})\right)  -\ln \left(  1-\sum_{k=1}^{l-1}\alpha(\mathbf{z}_{j}%
,\mathbf{x}_{k})\right)  \right]  .\label{Beran_est_ln_20}%
\end{equation}

The unconstrained optimization problem has been obtained, which can be solved
by means of a gradient-based algorithm. The weight is defined as
$w_{j}=K^{\ast}(\mathbf{x},\mathbf{z}_{j})$, where $K^{\ast}(\cdot,\cdot)$ is
a kernel function which measures how similar two feature vectors. For example,
if to use the Gaussian kernel, then the weight is defined as
\begin{equation}
w_{j}=\exp \left(  -\left \Vert \mathbf{x}-\mathbf{z}_{j}\right \Vert ^{2}%
/\sigma \right)  ,\label{Beran_est_22}%
\end{equation}
where $\sigma$ is the kernel parameter.

It should be noted that we use two different kernels. The first one is used
for computing weights $\alpha(\mathbf{x},\mathbf{x}_{i})$ in the Beran
estimator. The second kernel is used to find weights $w_{j}$ of the generated
points in a local area around the explained example. The kernels also have
different parameters. 

Finally, we write Algorithm \ref{alg:SurvBerLIME1} implementing the proposed
explanation method.

\begin{algorithm}
\caption{The algorithm implementing SurvBeX for computing vector $\bf {b}$ of the feature importance for point $\bf {x}$ }\label{alg:SurvBerLIME1}%

\begin{algorithmic}
[1]\REQUIRE Training set $\mathcal{A}$; point of interest $\mathbf{x}$; the
number of generated points $N$; parameters of kernels $\tau$ and $\sigma$.

\ENSURE Vector $\mathbf{b}$ of important features of $\mathbf{x}$.

\STATE Generate $N$ random points $\mathbf{z}_{k}$ in a local area around
$\mathbf{x}$.

\STATE Find predictions of SFs $S(t|\mathbf{z}_{j})$, $j=1,...,N$, produced by
the black-box survival model.

\STATE Compute weights $w_{j}=K^{\ast}(\mathbf{x},\mathbf{z}_{j})$ of
generated points, $j=1,...,N$, by using (\ref{Beran_est_22}).

\STATE Construct the optimization problem (\ref{Beran_est_16}) by using
(\ref{Beran_est_ln_20}) and (\ref{Beran_est}) with weights $\alpha
(\mathbf{x},\mathbf{x}_{i},\mathbf{b})$ expressed through (\ref{Beran_est_32}).

\STATE Find vector $\mathbf{b}$ by solving the optimization problem
(\ref{Beran_est_16}) using the gradient-based algorithm.
\end{algorithmic}
\end{algorithm}

If to use the logarithmic representation (\ref{Beran_est_18}), then the
computational difficulties of calculating the logarithm of the SF produced by
the Beran estimator can be simplified. The idea behind the simplification is
to use the series expansion of the logarithm as $\ln(1-x)\approx-x-x^{2}/2$.
Let us denote $\alpha(\mathbf{x},\mathbf{x}_{j})=\alpha_{j}$ for short. By
using the series expansion of the logarithm, we can rewrite the $i$-th term in
(\ref{Beran_est_ln_20}) as follows:%
\begin{align}
&  \ln \left(  1-\sum_{j=1}^{i}\alpha_{j}\right)  -\ln \left(  1-\sum
_{j=1}^{i-1}\alpha_{j}\right) \nonumber \\
&  \approx-\sum_{j=1}^{i}\alpha_{j}-\frac{1}{2}\left(  \sum_{j=1}^{i}%
\alpha_{j}\right)  ^{2}+\sum_{j=1}^{i-1}\alpha_{j}+\frac{1}{2}\left(
\sum_{j=1}^{i-1}\alpha_{j}\right)  ^{2}\nonumber \\
&  =-\alpha_{i}-\frac{1}{2}\left(  \sum_{j=1}^{i}\alpha_{j}-\sum_{j=1}%
^{i-1}\alpha_{j}\right)  \left(  \sum_{j=1}^{i}\alpha_{j}+\sum_{j=1}%
^{i-1}\alpha_{j}\right) \nonumber \\
&  =-\alpha_{i}-\frac{1}{2}\alpha_{i}\left(  \alpha_{i}+2\sum_{j=1}%
^{i-1}\alpha_{j}\right)  =-\alpha_{i}\left(  1-\frac{1}{2}\alpha_{i}%
+\sum_{j=1}^{i}\alpha_{j}\right)  .
\end{align}

Hence, we obtain
\begin{align}
\ln S_{B}(t|\mathbf{x},\mathbf{b})  &  \approx-\sum \nolimits_{t_{i}\leq
t}\delta_{i}\cdot \alpha(\mathbf{x},\mathbf{x}_{i})\nonumber \\
&  \times \left(  1-\frac{1}{2}\alpha(\mathbf{x},\mathbf{x}_{i})+\sum_{j=1}%
^{i}\alpha(\mathbf{x},\mathbf{x}_{j})\right)  .
\end{align}

At the same time, another problem of using the logarithm of the SF is its
extremely small negative values when $S_{B}$ is close to $0$. In order to
overcome this difficulty, the SF $S_{B}$ should be restricted by some positive
small value $\varepsilon$.

\section{Numerical experiments with synthetic data}

In order to study the proposed explanation algorithm, we generate random
survival times to events by using the Cox model estimates to have a
groundtruth vector of parameters $\mathbf{b}$. This allows us to compare
different explanation methods. 

As black-box models, we use the Cox model, the RSF \cite{Ibrahim-etal-2008}
and the Beran estimator. The RSF consists of $100$ decision survival trees.
The number of features selected to build each tree in RSF is $\sqrt{d}$. The
largest depth of each tree is $8$. The log-rank splitting rule is used. The
Gaussian kernels with the parameter (the width), which is equal to $1/250$, is
used in the Beran estimator.

The main reason for using the Cox model as a black box is to check whether the
selected important features explaining the SF at the explained point
$\mathbf{x}$ coincide with the corresponding features accepted in the Cox
model for generating training set. It should be noted that the Cox model as
well as the RSF are viewed as black-box models whose predictions (CHFs or
survival functions) are explained.

The well-known Broyden--Fletcher--Goldfarb--Shanno (BFGS) algorithm
\cite{Broyden-1970,Fletcher-1970,Goldfarb-1970,Shanno-1970} is used to solve
the optimization problem (\ref{Beran_est_16}).

Two kernels as used in the Beran estimator incorporated into SurvBeX: the
standard Gaussian with the parameter $\tau$ taking values $10$ or $100$ and
the kernel of the form:
\[
K\left(  \mathbf{x},\mathbf{x}_{i}\right)  =\exp \left(  \frac{\left \vert
\mathbf{x}-\mathbf{x}_{i}\right \vert }{\tau}\right)  .
\]

\subsection{Generation of random covariates, survival times and perturbations}

In order to evaluate SurvBeX and to compare it with SurvLIME, we use synthetic
data. Two types of generated datasets are studied. The first one contains
examples which are concentrated around one center. Each feature of a point
$\mathbf{x}\in \mathbb{R}^{d}$ is randomly generated from the uniform
distribution around the center $p$ with radius $R.$Random survival times in
accordance with the Cox model are generated by applying the method proposed in
\cite{Bender-etal-2005}. According to \cite{Bender-etal-2005}, generated
survival times have the Weibull distribution with the scale $\lambda$ and
shape $v$ parameters. The use of the Weibull distribution is justified by the
fact that the Weibull distribution supports the assumption of proportional
hazards \cite{Bender-etal-2005}. If we denote the random variable uniformly
distributed in interval $[0,1]$ as $U$, then the random survival time is
generated as follows \cite{Bender-etal-2005}:%
\begin{equation}
T=\left(  \frac{-\ln(U)}{\lambda \exp(\mathbf{b}^{true}\mathbf{x}^{\mathrm{T}%
})}\right)  ^{1/v}.
\end{equation}

Here the vector $\mathbf{b}^{true}=(b_{1}^{true},...,b_{d}^{true})$ is the
vector of known coefficients that are set in advance to generate data.

Numerical experiments are performed under condition of different numbers $d$
of features, which are $d=5$, $d=10$, and $d=20$.

Parameters of the generation for every number of features are the following:

\begin{enumerate}
\item $d=5$: $\mathbf{b}^{true}$ $=(0.5,0.25,0.12,0,0);$

\item $d=10$: $\mathbf{b}^{true}$ $=(0.6,0.3,0.1,0,0,...,0)$;

\item $d=20$: $\mathbf{b}^{true}$ $=(0.5,0.25,0.12,0,0,...,0)$;
\end{enumerate}

The vectors $\mathbf{b}^{true}$ are chosen in such a way that the feature
importance values differ significantly. This will make it easier to compare
the feature importance for different models. Parameters of the Weibull
distribution for generating survival times are $\lambda=10^{-5}$, $v=2$. The
number of generated points $(\mathbf{x}_{i},\delta_{i},T_{i})$ in the dataset
is $n=200$. The center of the single cluster is $p=(0.5,0.5,0.5,0.5)$, the
radius is $R=0.5$.

\begin{figure}
[ptb]
\begin{center}
\includegraphics[
height=1.8011in,
width=3.2356in
]%
{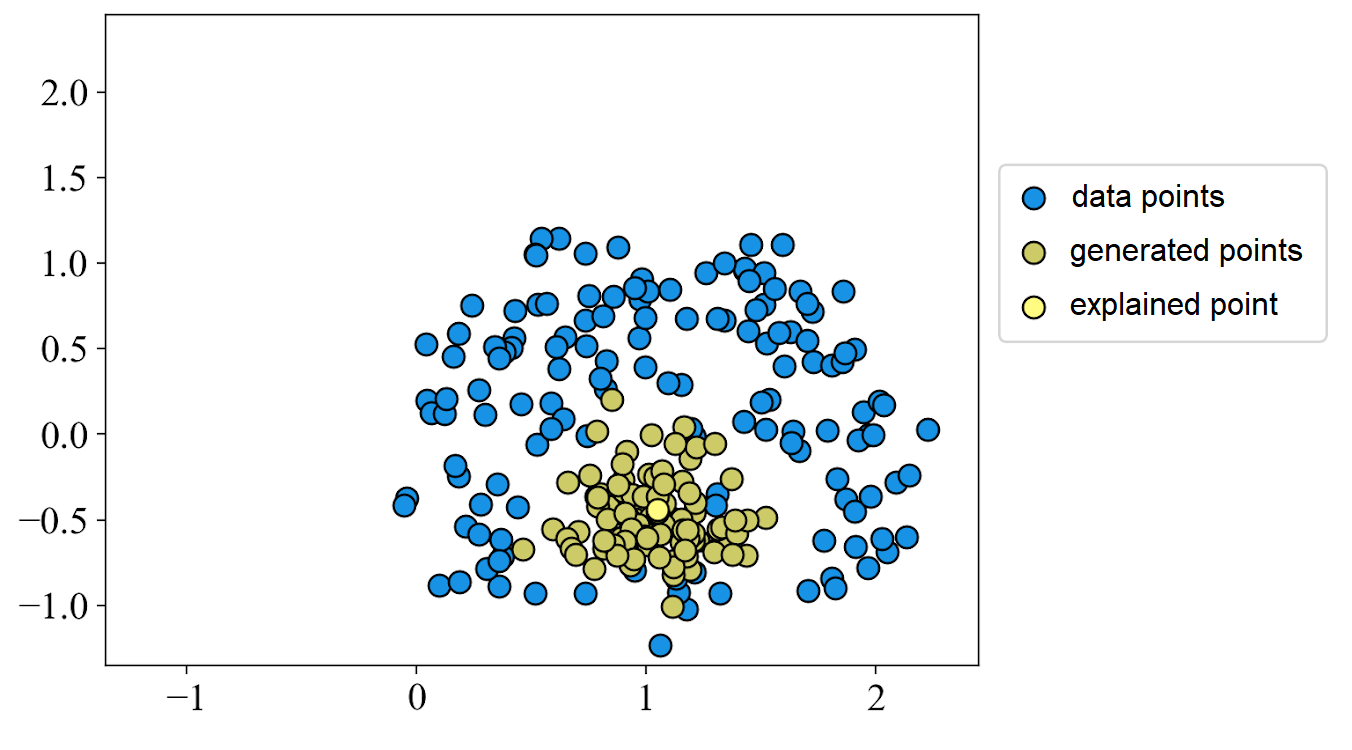}%
\caption{An example of the one-cluster data structure with the dataset points,
an explained point, and points generated around the explained point }%
\label{f:one_cluster}%
\end{center}
\end{figure}

The number of generated points around an explained example $\mathbf{z}$ is
$N=100$. These points are generated in accordance with the normal distribution
with the expectation $z$ (the explained example) with the standard deviation
$0.4\cdot0.5$. Weights $w_{j}$ are assigned to every generated point by using
the Gaussian kernel with the parameter $\sigma=0.4$. An example of the
one-cluster data structure generated in accordance with the above parameters
by using the t-SNE method is depicted in Fig.\ref{f:one_cluster}.%

The second type of generated datasets consists of two clusters with different
centers $p^{(1)}$ and $p^{(2)}$ of feature vectors and different vectors of
the Cox model coefficients $\mathbf{b}_{1}^{true}$ and $\mathbf{b}_{2}^{true}%
$. The reason for using such a two-cluster data structure is to try to
complicate the Cox model data structure. If examples are distributed according
to the Cox model, then it is likely that the Cox explanation model underlying
SurvLIME will provide a more accurate explanation than SurvBeX. Therefore, it
is proposed to use two data clusters for a correct comparison of the models.

Parameters of the cluster generation also depend on the number of features
$d$. For the numbers of features $d=5$, $d=10$, and $d=20$, they are

\begin{enumerate}
\item $d=5$:

\begin{enumerate}
\item $\mathbf{b}_{1}^{true}=(0.5,0.25,0.12,0,0)$;

\item $\mathbf{b}_{2}^{true}$ $=(0,0,0.12,0.25,0.5)$;
\end{enumerate}

\item $d=10$:

\begin{enumerate}
\item $\mathbf{b}_{1}^{true}$ $=(0.6,0.3,0.1,0,0,...,0)$;

\item $\mathbf{b}_{2}^{true}$ $=(0,0,...,0,0.1,0.3,0.6);$
\end{enumerate}

\item $d=20$:

\begin{enumerate}
\item $\mathbf{b}_{1}^{true}$ $=(0.5,0.25,0.12,0,0,...,0)$;

\item $\mathbf{b}_{2}^{true}$ $=(0,0,...,0,0.12,0.25,0.5).$
\end{enumerate}
\end{enumerate}

The following parameters for points of every cluster are used:

\begin{enumerate}
\item cluster 1: center $p^{(1)}=(0.25,0.25,0.25,0.25,0.25)$; radius $R=0.2$;
the number of points in the cluster $n=200$;

\item cluster 2: center $p^{(2)}=(0.75,0.75,0.75,0.75)$; radius $R=0.2$; the
number of points in the cluster $n=200$.
\end{enumerate}

\begin{figure}
[ptb]
\begin{center}
\includegraphics[
height=1.9841in,
width=3.8257in
]%
{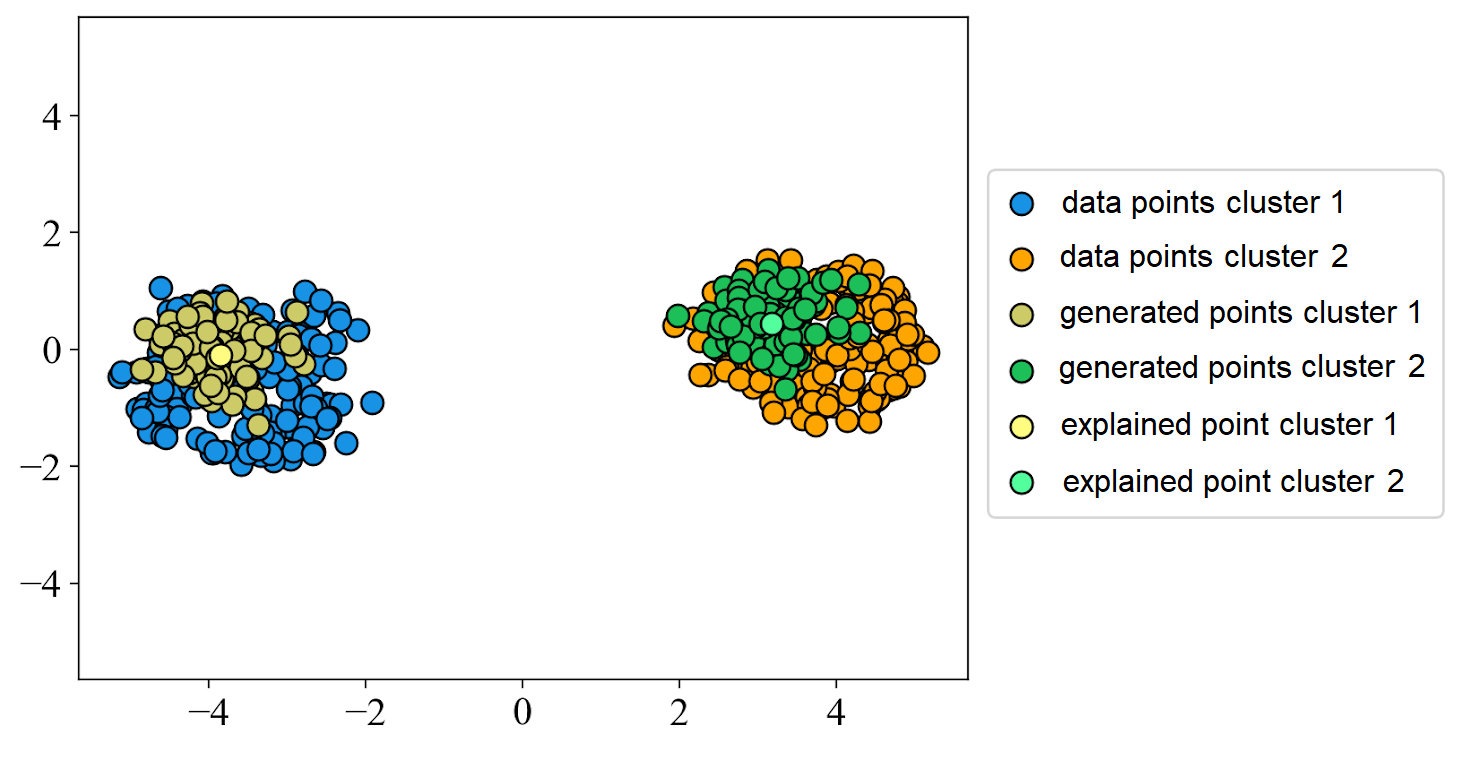}%
\caption{An example of the two-cluster data structure with the dataset points,
an explained point, and points generated around the explained point in each
cluster }%
\label{f:two_clusters}%
\end{center}
\end{figure}

Parameters of the Weibull distribution for generating survival times are
$\lambda=10^{-5}$, $v=2$. The total number of generated points $(\mathbf{x}%
_{i},\delta_{i},T_{i})$ in the dataset consisting of two clusters is $n=400$.
The total number of generated points around an explained example is $N=100$.
Parameters of the generation are the same as in experiments with one cluster.
An example of the two-cluster data structure generated in accordance with the
above parameters by using the t-SNE method is depicted in
Fig.\ref{f:two_clusters}.%

\subsection{Measures for comparison the models}

To compare different models (SurvLIME and SurvBeX), we use the following three
measures:
\[
D(\mathbf{b}^{{model}},\mathbf{b}^{true})=\left \Vert \mathbf{b}^{{model}%
}-\mathbf{b}^{true}\right \Vert ^{2},
\]%
\[
KL(\mathbf{b}^{{model}},\mathbf{b}^{true})=\sum_{i=1}^{n}b_{i}^{true}\ln
\frac{b_{i}^{true}}{b_{i}^{{model}}},
\]%
\[
C(\mathbf{b}^{{model}},\mathbf{b}^{true})=\frac{\sum \nolimits_{i,j}%
\mathbf{1}[b_{i}^{true}<b_{j}^{true}]\cdot \mathbf{1}[b_{i}^{{model}}%
<b_{j}^{{model}}]}{\sum \nolimits_{i,j}\mathbf{1}[b_{i}^{true}<b_{j}^{true}]}.
\]

The first measure represents the Euclidean distance between the two vectors
$\mathbf{b}^{{model}}$ and $\mathbf{b}^{true}$, where the vector
$\mathbf{b}^{{model}}$ is obtained by SurvLIME and SurvBeX. The vectors
$\mathbf{b}^{{model}}$ and $\mathbf{b}^{true}$ are normalized. Therefore, they
can be regarded as the probability distributions and the Kullback--Leibler
divergence (KL) can be applied to analyze the difference between the vectors,
which is used as the second measure. The third measure is the C-index applied
to the vectors $\mathbf{b}^{{model}}$ and $\mathbf{b}^{true}$. It estimates
the probability that any pair of vectors $\mathbf{b}^{{model}}$ and
$\mathbf{b}^{true}$ are correctly ranking. It should be noted that the third
index is the most informative because it estimates the relationship between
elements of two vectors. The Euclidean distance and the KL-divergence show how
the vectors are close to each other, but they do not show the relationship
between elements of two vectors. Nevertheless, these measures can be also
useful when the models are compared.

In order to study how SFs predicted by using different methods, including the
black-box, SurvLIME, and SurvBeX, are close to each other, we use the
following distance measure:
\begin{equation}
D(S,S_{B-B})=\sum_{i=1}^{n}\left(  S^{(i)}(\mathbf{z})-S_{B-B}^{(i)}%
(\mathbf{z})\right)  ^{2}\left(  t_{i}-t_{i-1}\right)  ,
\end{equation}
where $S^{(i)}(\mathbf{z})$ is the SF predicted by SurvLIME or SurvBeX for the
explained point $\mathbf{z}$ in the time interval $[t_{i-1},t_{i}]$;
$S_{B-B}^{(i)}$ is the SF predicted by the black-box model in the same time interval.

When $M$ points are used for explanation, then the mean values of the above
measures are computed: mean squared distance (MSD), which can be viewed as an
analog of the mean squared error (MSE), mean KL-divergence (MKL), mean C-index
(MCI), and mean SF distance (MSFD). The above measures are computed similarly,
for example, the MSD is determined as
\[
MSD=\frac{1}{M}\sum_{i=1}^{M}D(\mathbf{b}_{i}^{{model}},\mathbf{b}^{true}),
\]
where $\mathbf{b}_{i}^{{model}}$ is the vector obtained by solving
(\ref{Beran_est_18}) for the $i$-th explained point by using SurvLIME or SurvBeX.

The greater the value of the MCI and the smaller the MSFD, the MKL, and the
MSD, the better results we get.

\subsection{One cluster}

First, we compare SurvBeX with SurvLIME by generating one cluster as it is
shown in Fig.\ref{f:one_cluster}. The first black-box model is the Cox model
which is used in order to be sure that SurvLIME provides correct results. It
is important to point out the following observation. If the Cox model uses the
same method for computing the baseline SF $S_{0}(t\mathbf{)}$ (see the
corresponding expression (\ref{Cox_SF})), then SurvLIME provides outperforming
results because it actually tries to minimize the distance between SFs which
are similar to some extent. In this case, we obtain outperforming results
provided by SurvLIME. The outperformance is also supported by the fact that
the training set is generated in accordance with probability distributions
satisfying the Cox model.

\begin{figure}
[ptb]
\begin{center}
\includegraphics[
height=3.9277in,
width=4.1027in
]%
{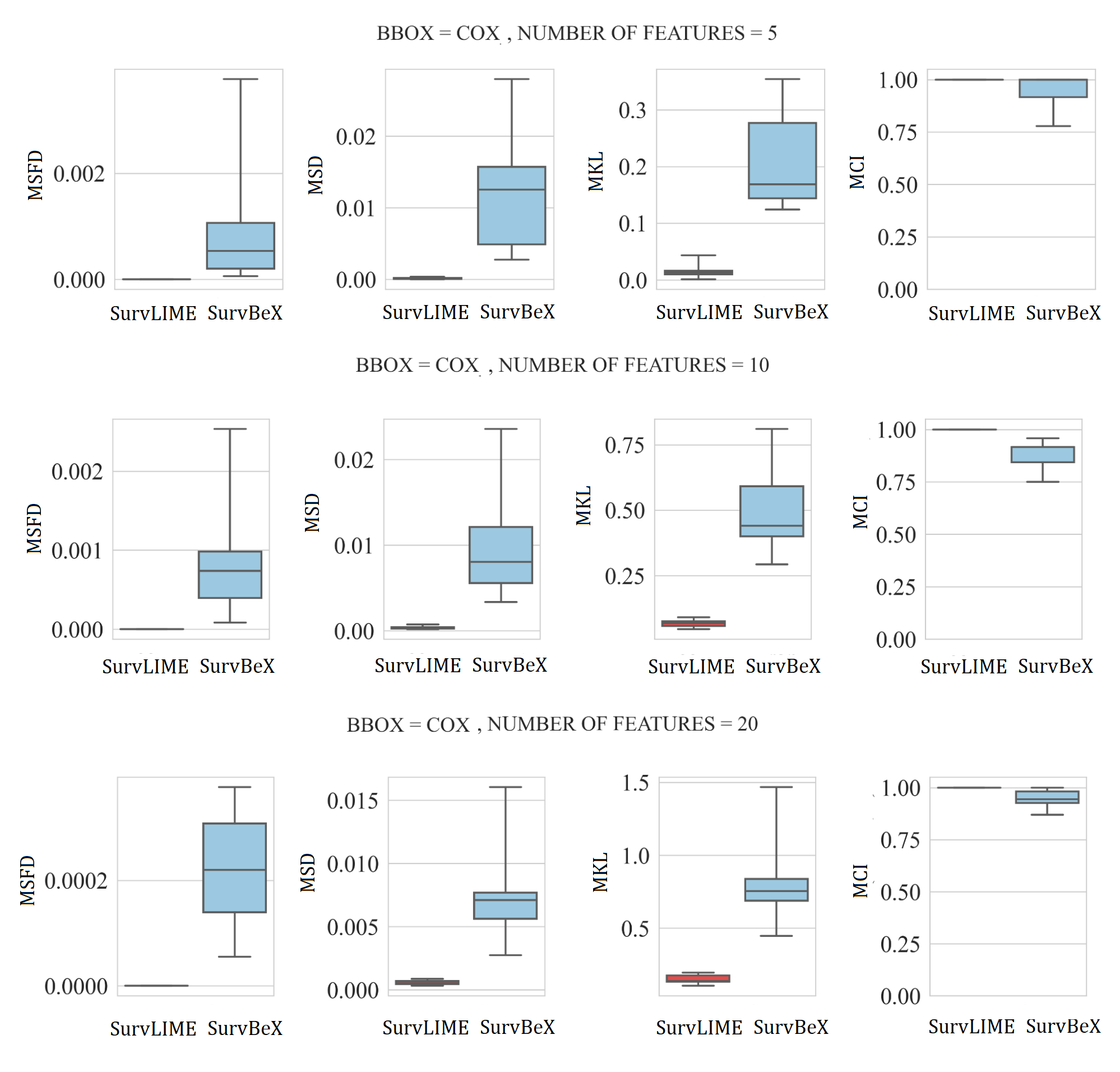}%
\caption{Boxplots illustrating difference between MSFD, MSD, MKL, and MCI for
SurvLIME and SurvBeX when the black box is the Cox model and numbers of
features are 5, 10, 20}%
\label{f:cox_c1_m}%
\end{center}
\end{figure}

Boxplots in Fig.\ref{f:cox_c1_m} show differences between MSFD, MSD, MKL, and
MCI for SurvLIME and SurvBeX when the black box is the Cox model. It can be
seen from Fig.\ref{f:cox_c1_m} that SurvLIME provides better results in
comparison with SurvBeX in accordance with all introduced measures. The
results justify the above assumption about comparison of SurvLIME and SurvBeX
when the black box is the Cox model and training example are generated in
accordance with the Cox model.%

\begin{figure}
[ptb]
\begin{center}
\includegraphics[
height=4.2707in,
width=4.064in
]%
{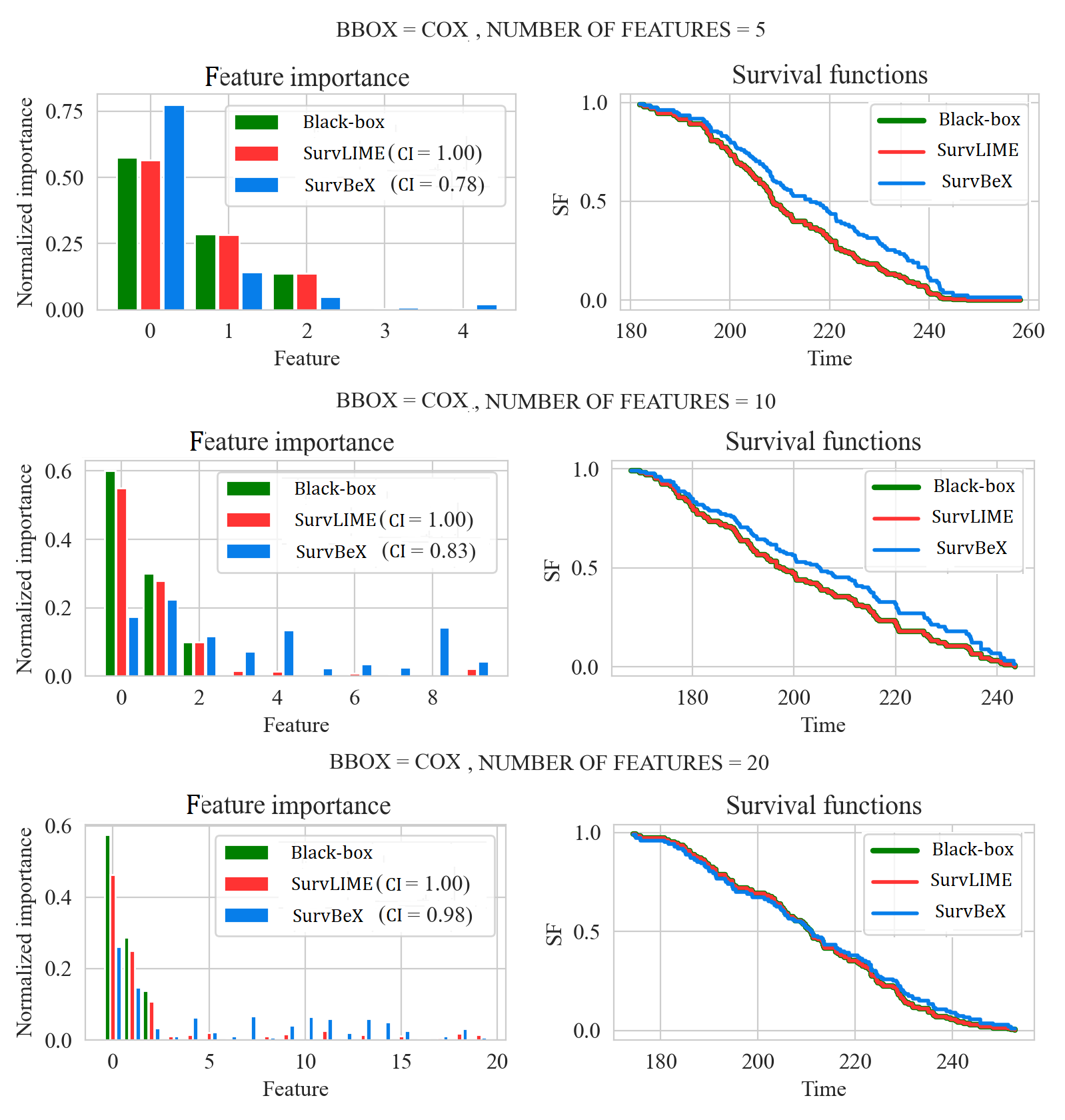}%
\caption{Values of the feature importance (the left column) and SFs provided
by the black-box model (the Cox model), SurvLIME, and SurvBeX (the right
column)}%
\label{f:cox_c1_sf}%
\end{center}
\end{figure}

Fig.\ref{f:cox_c1_sf} shows values of the feature importance (the left column)
and SFs provided by the black box, SurvLIME and SurvBeX (the right column)
under the same conditions. For each number of features in examples of the
generated dataset, an explained example is randomly selected from the dataset.
Results of the explanation and the corresponding SFs provided by different
models are depicted. It can be seen from Fig.\ref{f:cox_c1_sf} that SFs
produced by the black-box model (depicted by the thicker line) and SurvLIME
almost coincide whereas SF produced by SurvBeX differs from the black-box
model SF. At the same time, when the number of features is 20, SurvBeX
provides results similar to SurvLIME.%

\begin{figure}
[ptb]
\begin{center}
\includegraphics[
height=3.9277in,
width=4.3516in
]%
{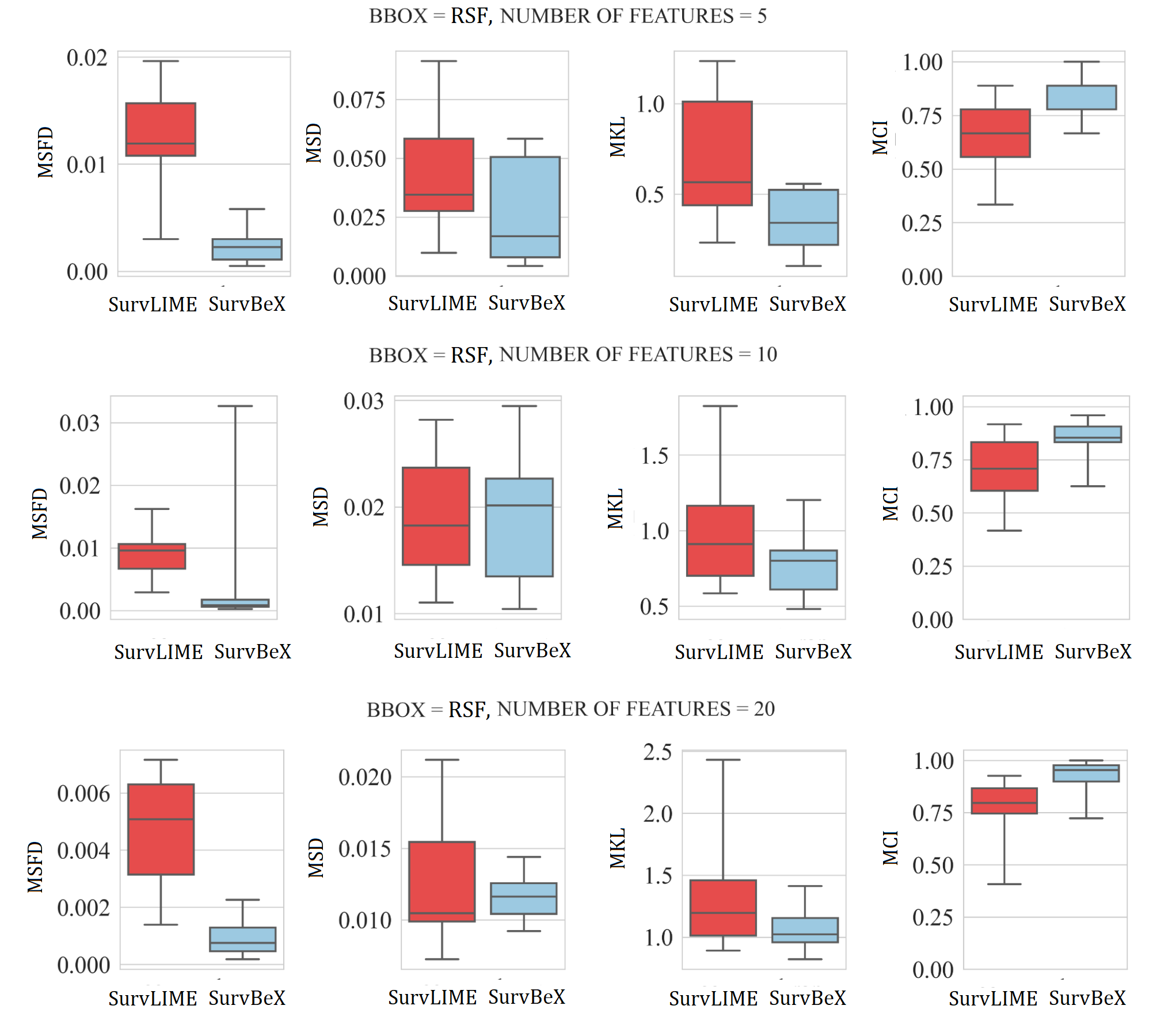}%
\caption{Boxplots illustrating difference between MSFD, MSD, MKL, and MCI for
SurvLIME and SurvBeX when the black box is the RSF and numbers of features are
5, 10, 20}%
\label{f:rf_c1_m}%
\end{center}
\end{figure}

\begin{figure}
[ptb]
\begin{center}
\includegraphics[
height=4.0922in,
width=3.918in
]%
{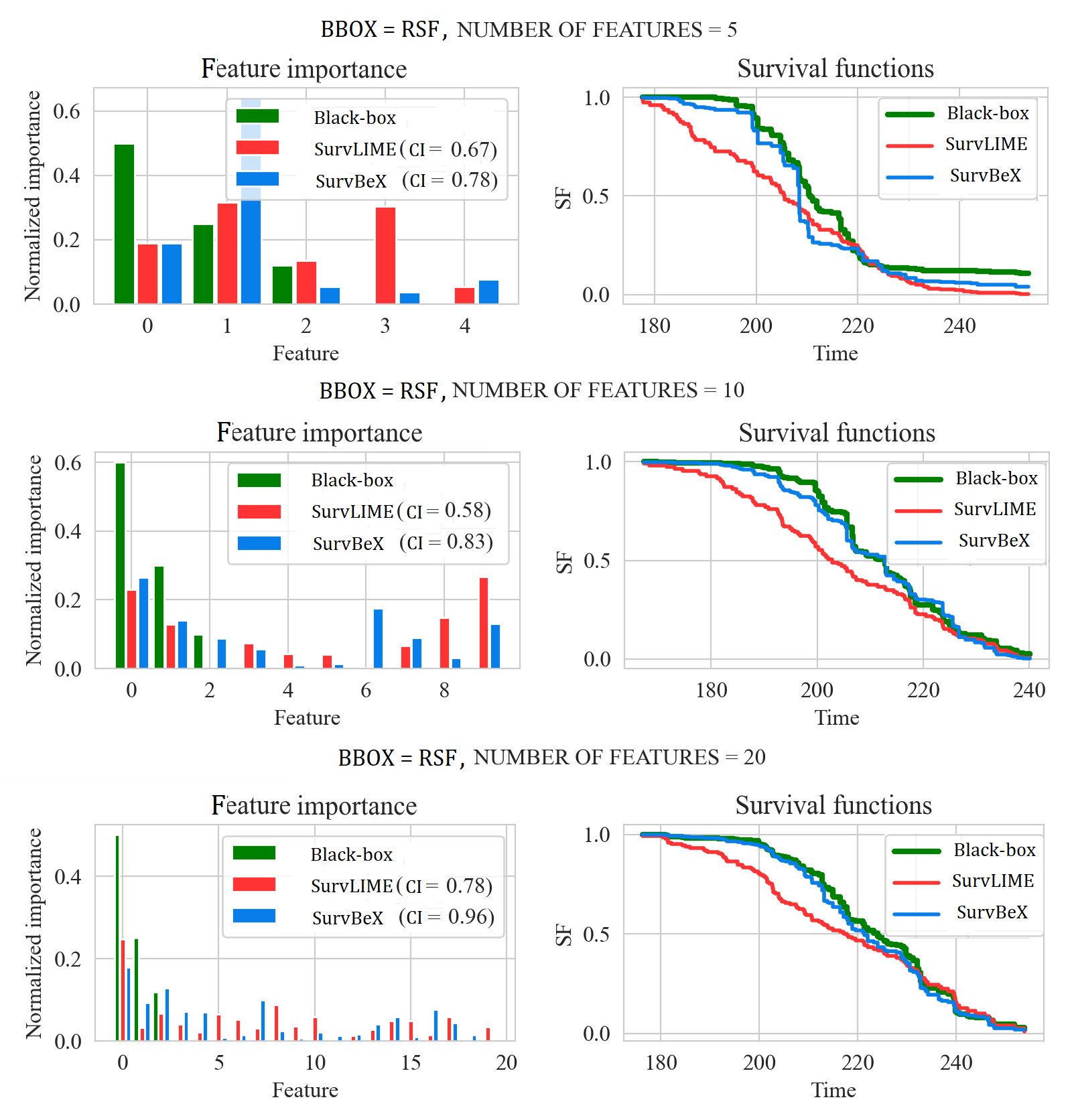}%
\caption{Values of the feature importance (the left column) and SFs provided
by the black-box model (the RSF), SurvLIME, and SurvBeX (the right column)}%
\label{f:rf_c1_sf}%
\end{center}
\end{figure}

Let us consider the RSF as a black-box model. The corresponding numerical
results are shown in Figs.\ref{f:rf_c1_m}-\ref{f:rf_c1_sf}.%

Fig.\ref{f:rf_c1_m} shows boxplots of MSFD, MSD, MKL, and MCI for SurvLIME and
SurvBeX when the black box is the RSF. It is interesting to note that SurvBeX
provides better results in comparison with SurvLIME. It is interesting also to
point out that SurvBeX behaves more stable when the number of features
increases. It can be also seen from Fig.\ref{f:rf_c1_m} (boxplots of MSFD)
that the SurvBeX provides better approximation of the SF because the MSFD
measures for SurvBeX are smaller than the same measures for SurvLIME. The same
conclusion is supported by Fig.\ref{f:rf_c1_sf} where randomly selected
explained examples for three cases of the feature numbers are analyzed. It can
be seen from Fig.\ref{f:rf_c1_sf} that SFs provided by SurvBeX are much closer
to the black-box SF in comparison with SFs produced by SurvLIME. The
difference between results obtained by different black-box models can be
explained by the fact that, in spite of the training set generated by using
the Cox model, the RSF produces the SF different from the \textquotedblleft
ideal\textquotedblright \ SF corresponding to the Cox model and this difference
increases with the number of features. The Beran estimator is more sensitive
to the changes of SFs and, therefore, SurvBeX provides better results.%

\begin{figure}
[ptb]
\begin{center}
\includegraphics[
height=3.6076in,
width=4.7966in
]%
{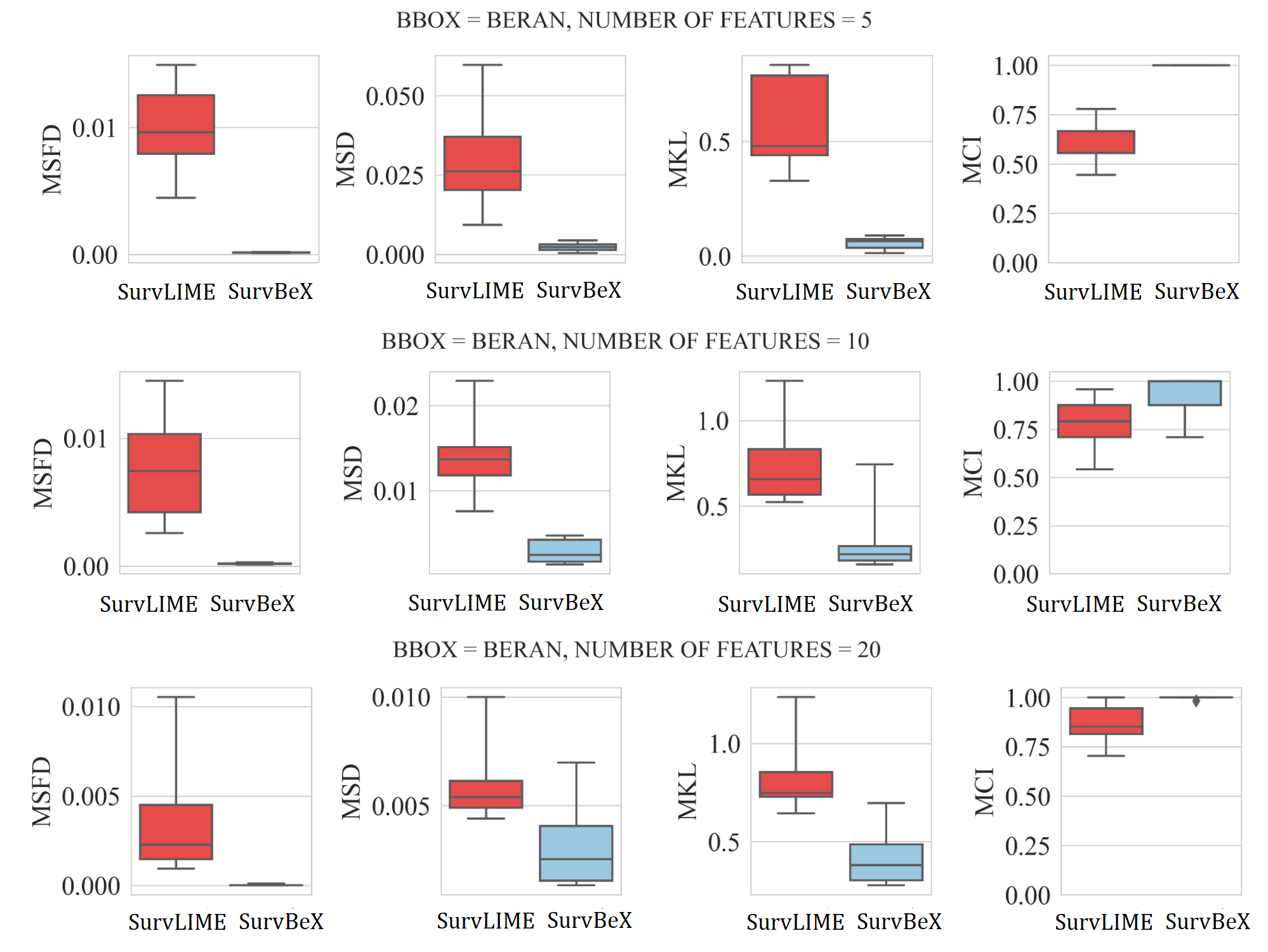}%
\caption{Boxplots illustrating difference between MSFD, MSD, MKL, and MCI for
SurvLIME and SurvBeX when the black box is the Beran estimator}%
\label{f:ber_c1_m}%
\end{center}
\end{figure}

\begin{figure}
[ptb]
\begin{center}
\includegraphics[
height=4.232in,
width=3.9937in
]%
{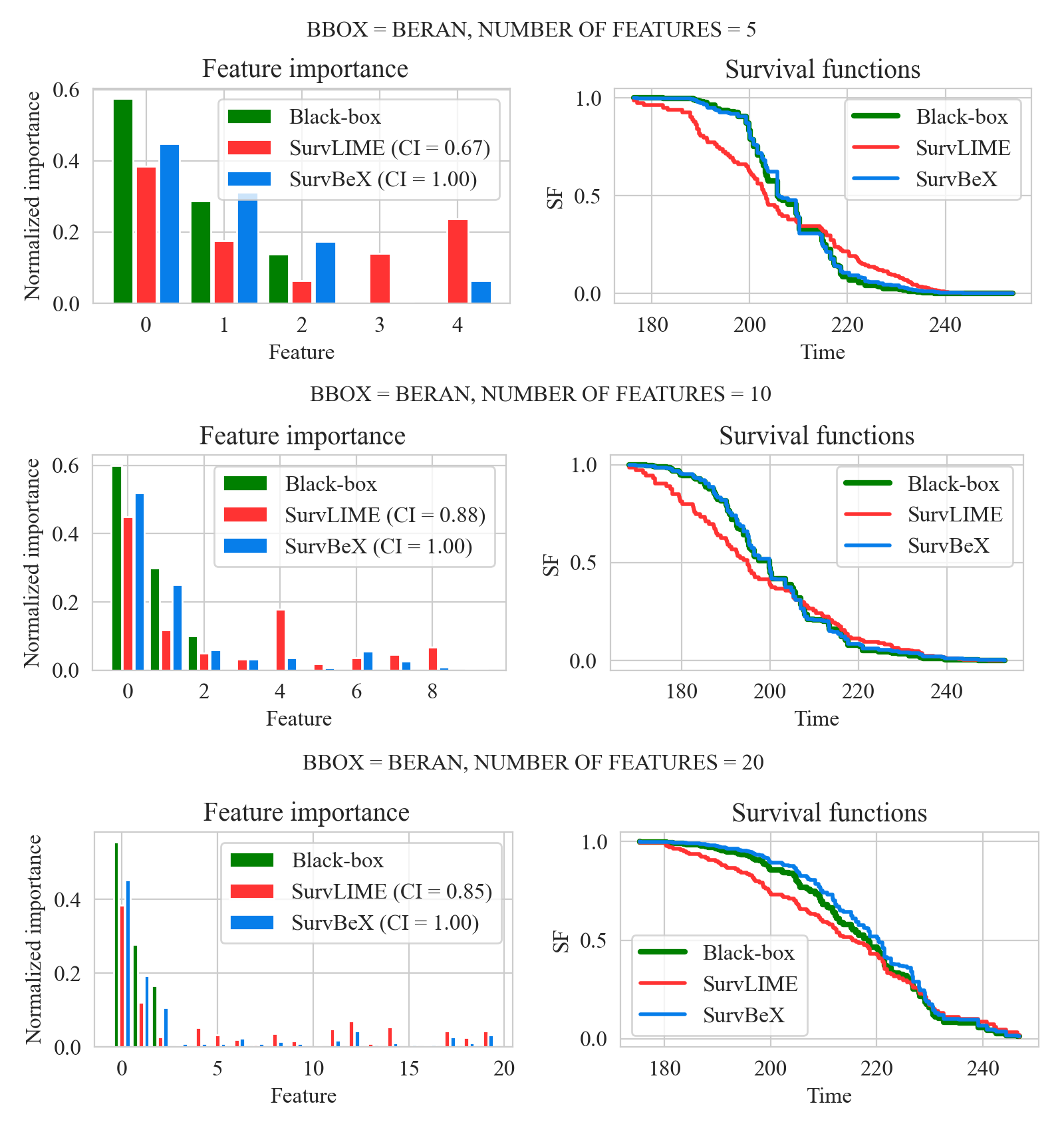}%
\caption{Values of the feature importance (the left column) and SFs provided
by the Beran estimator as a black-box model, SurvLIME, and SurvBeX (the
right column)}%
\label{f:ber_c1_sf}%
\end{center}
\end{figure}

In order to study different cases of the black-box models, we use the Beran
estimator as a black-box model. It is obvious in this case that SurvBeX should
provide outperforming results. This assumption is justified by results shown
in Fig.\ref{f:ber_c1_m}. One can see from Fig.\ref{f:ber_c1_m} that SurvBeX
outperforms SurvLIME for all numbers of features. The same can be seen from
Fig.\ref{f:ber_c1_sf}. SFs produced by SurvBeX are closer to the black-box SFs
than SFs produced by SurvLIME.%

%

\subsection{Two clusters}

Let us compare SurvBeX with SurvLIME when two clusters of examples are
generated (see Fig.\ref{f:two_clusters}) such that the clusters are generated
in accordance with the Cox model, but they have quite different parameters of
the model. It is obvious that this data structure entirely violates the
homogeneous Cox model data structure. Therefore, it can be assumed in advance
that SurvLIME will produce worse results. This is confirmed by the following experiments.

\begin{figure}
[ptb]
\begin{center}
\includegraphics[
height=3.8204in,
width=4.4387in
]%
{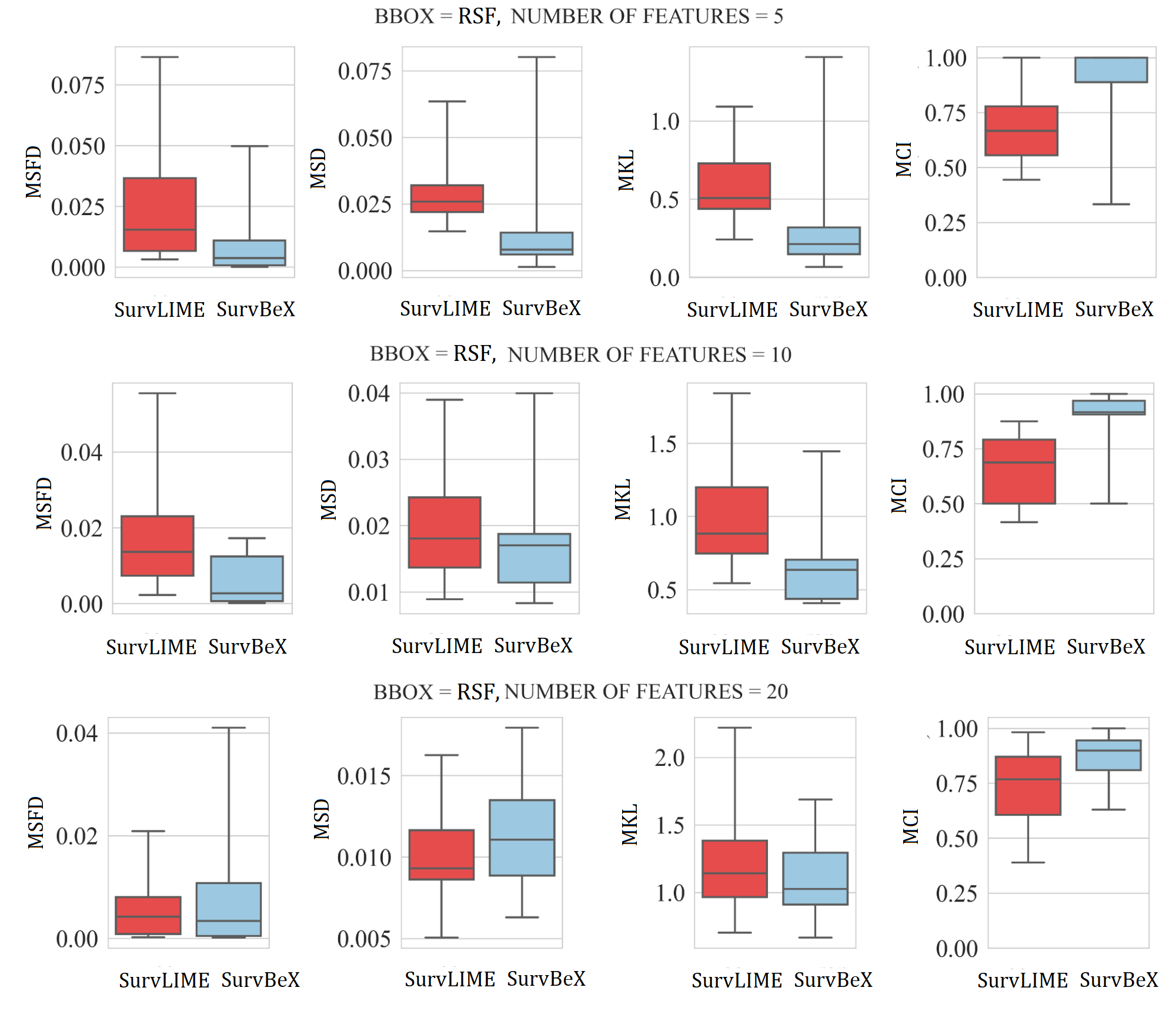}%
\caption{Boxplots illustrating difference between MSFD, MSD, MKL, and MCI for
SurvLIME and SurvBeX when the black box is the RSF and two clusters of
training data are used}%
\label{f:rf_c2_m}%
\end{center}
\end{figure}

We consider the case when the RSF is used as a black box. Fig.\ref{f:rf_c2_m}
shows boxplots of MSFD, MSD, MKL, and MCI for SurvLIME and SurvBeX. It can be
seen from Fig.\ref{f:rf_c2_m} that SurvBeX provides better results for
different numbers of features. At the same time, the outperformance of SurvBeX
is reduced with increase of the feature numbers.%

%

\begin{figure}
[ptb]
\begin{center}
\includegraphics[
height=4.1889in,
width=4.0297in
]%
{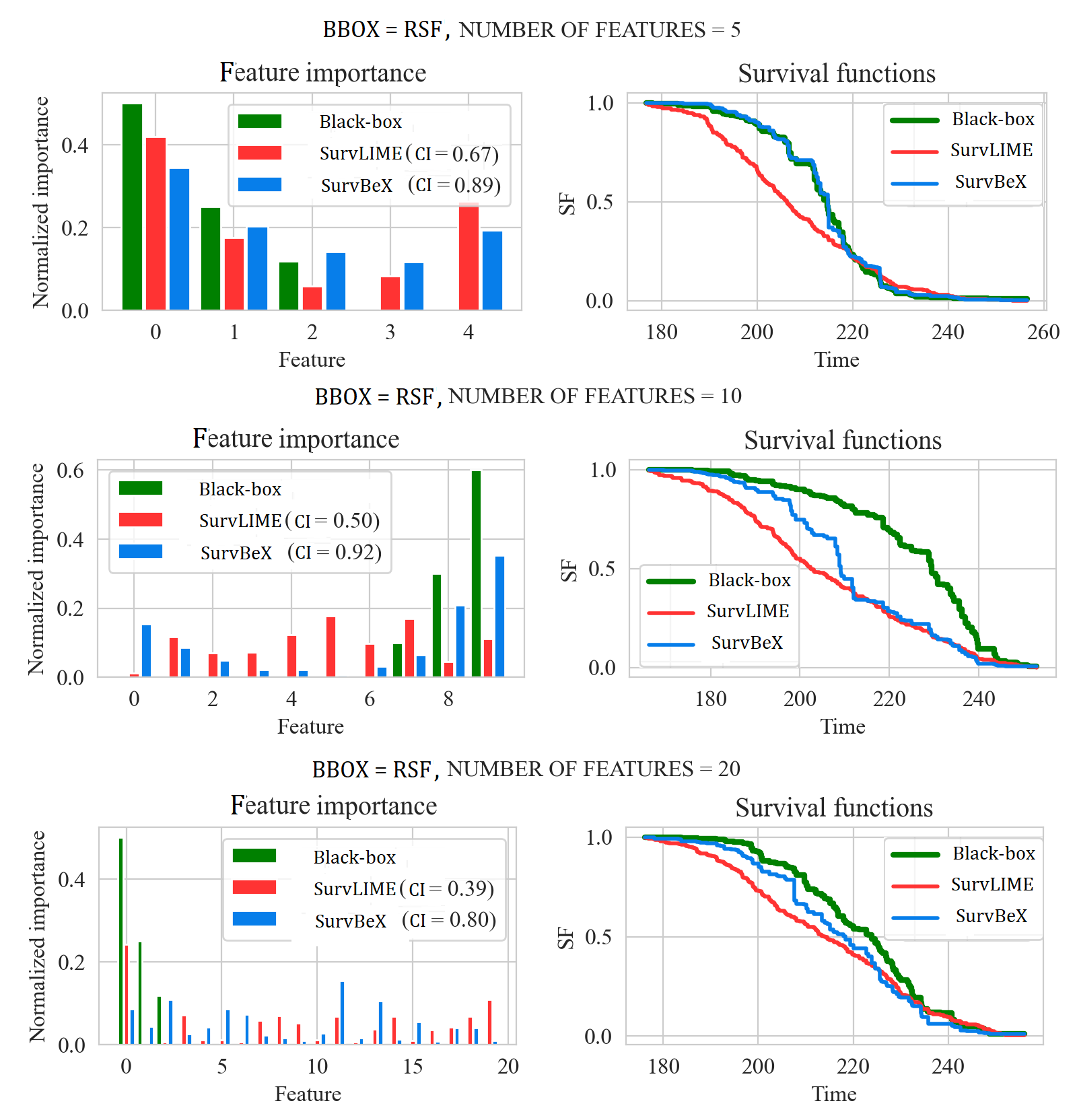}%
\caption{Values of the feature importance (the left column) and SFs provided
by the RSF as a black-box model, SurvLIME, and SurvBeX (the
right column) when two clusters are used}
\label{f:cox_c2_m}
\end{center}
\end{figure}

An important peculiarity of the RSF as a black-box model is that it tries to
separate examples from different clusters. As a result, the produced SF for
each point strongly depends on a cluster which contains the point. In contrast
to the RSF, the Cox model computes the baseline SF $S_{0}(t)$ which is common
for all points of clusters. Therefore, the Cox model as a black-box model
\textquotedblleft averages\textquotedblright \ the clusters and provides the
unsatisfactory prediction results. In particular, C-indices of the Cox
black-box model computed on the training and testing sets of the two-cluster
dataset with five features are $0.59$ and $0.57$, respectively, whereas
C-indices of the black-box RSF are $0.95$ and $0.91$, respectively. We can see
that the Cox black-box model is no better than random guessing.
Fig.\ref{f:cox_c2_m} shows boxplots of MSFD, MSD, MKL, and MCI for this case.
It seems from Fig.\ref{f:cox_c2_m} that SurvBeX is worse than SurvLIME.
However, this is not the case. The average distance (MSFD) between SFs is very
small, and explanation models simply try to guess the feature importance which
is also close to $0.5$. Due to the more complex optimization in SurvBeX, the
error of its explanation is greater.%

\begin{figure}
[ptb]
\begin{center}
\includegraphics[
height=1.5144in,
width=5.2777in
]%
{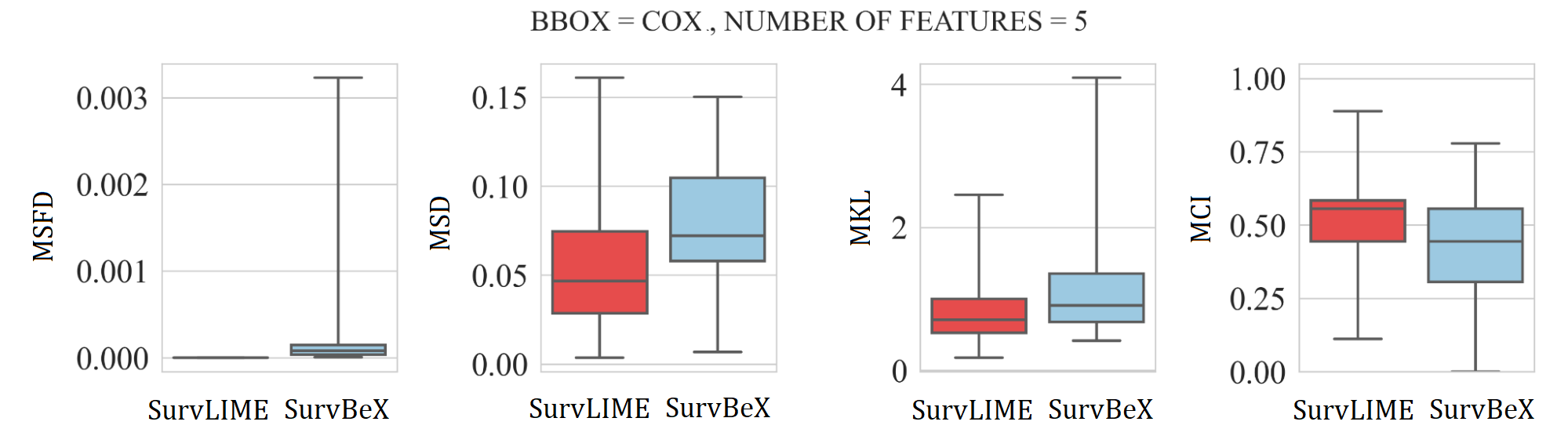}%
\caption{Boxplots illustrating difference between MSFD, MSD, MKL, and MCI for
SurvLIME and SurvBeX when the black box is the Cox model and the training
dataset consists of two clusters}%
\label{f:cox_c2_m}%
\end{center}
\end{figure}

Table \ref{t:summarize5} illustrate the relationship between different
explanation methods when two clusters are considered, the black-box is the
RSF, and the number of features in training examples is 5. We also consider an
additional explanation method called SurvSHAP \cite{Krzyzinski-etal-23}. Mean
values and standard deviations of measures MSD, MKL, and MCI are given in
Table \ref{t:summarize5}. It can be seen from the table that SurvBeX
outperforms other explanation methods. Similar results by $10$ features are
shown in Table \ref{t:summarize10}.%

\begin{table}[tbp] \centering
\caption{Mean values and standard deviations of measures MSD, MKL, and MCI for SurvBex, SurvLIME, and SurvSHAP by 5 features}%
\begin{tabular}
[c]{cccc}\hline
Method & MSD & MKL & MCI\\ \hline
SurvBeX & $0.004\pm0.002$ & $0.129\pm0.052$ & $0.956\pm0.057$\\
SurvLIME & $0.032\pm0.017$ & $0.621\pm0.343$ & $0.685\pm0.176$\\
SurvSHAP & $0.044\pm0.005$ & $0.629\pm0.069$ & $0.581\pm0.214$\\ \hline
\end{tabular}
\label{t:summarize5}%
\end{table}%
%

\begin{table}[tbp] \centering
\caption{Mean values and standard deviations of measures MSD, MKL, and MCI for SurvBex, SurvLIME, and SurvSHAP by 10  features}%
\begin{tabular}
[c]{cccc}\hline
Method & MSD & MKL & MCI\\ \hline
SurvBeX & $0.016\pm0.004$ & $0.665\pm0.151$ & $0.825\pm0.140$\\
SurvLIME & $0.016\pm0.005$ & $0.862\pm0.215$ & $0.642\pm0.139$\\
SurvSHAP & $0.035\pm0.002$ & $1.347\pm0.107$ & $0.583\pm0.192$\\ \hline
\end{tabular}
\label{t:summarize10}%
\end{table}%

\section{Numerical experiments with real data}

To illustrate SurvBeX, we study it on the following well-known real datasets:

\begin{itemize}
\item The \emph{Veterans' Administration Lung Cancer Study (Veteran) Dataset}
\cite{Kalbfleisch-Prentice-1980} contains data on 137 males with advanced
inoperable lung cancer. The subjects were randomly assigned to either a
standard chemotherapy treatment or a test chemotherapy treatment. Several
additional variables were also measured on the subjects. The dataset can be
obtained via the \textquotedblleft survival\textquotedblright \ R package or
the Python \textquotedblleft scikit-survival\textquotedblright \ package..

\item The \emph{German Breast Cancer Study Group 2 (GBSG2) Dataset} contains
observations of 686 women \cite{Sauerbrei-Royston-1999}. Every example is
characterized by 10 features, including age of the patients in years,
menopausal status, tumor size, tumor grade, number of positive nodes, hormonal
therapy, progesterone receptor, estrogen receptor, recurrence free survival
time, censoring indicator (0 - censored, 1 - event). The dataset can be
obtained via the \textquotedblleft TH.data\textquotedblright \ R package or the
Python \textquotedblleft scikit-survival\textquotedblright \ package.

\item The \emph{Worcester Heart Attack Study (WHAS500) Dataset}
\cite{Hosmer-Lemeshow-May-2008} describes factors associated with acute
myocardial infarction. It considers 500 patients with 14 features. The
endpoint is death, which occurred for 215 patients (43.0\%). The dataset can
be obtained via the \textquotedblleft smoothHR\textquotedblright \ R package or
the Python \textquotedblleft scikit-survival\textquotedblright \ package.`
\end{itemize}

\begin{figure}
[ptb]
\begin{center}
\includegraphics[
height=1.977in,
width=5.9645in
]%
{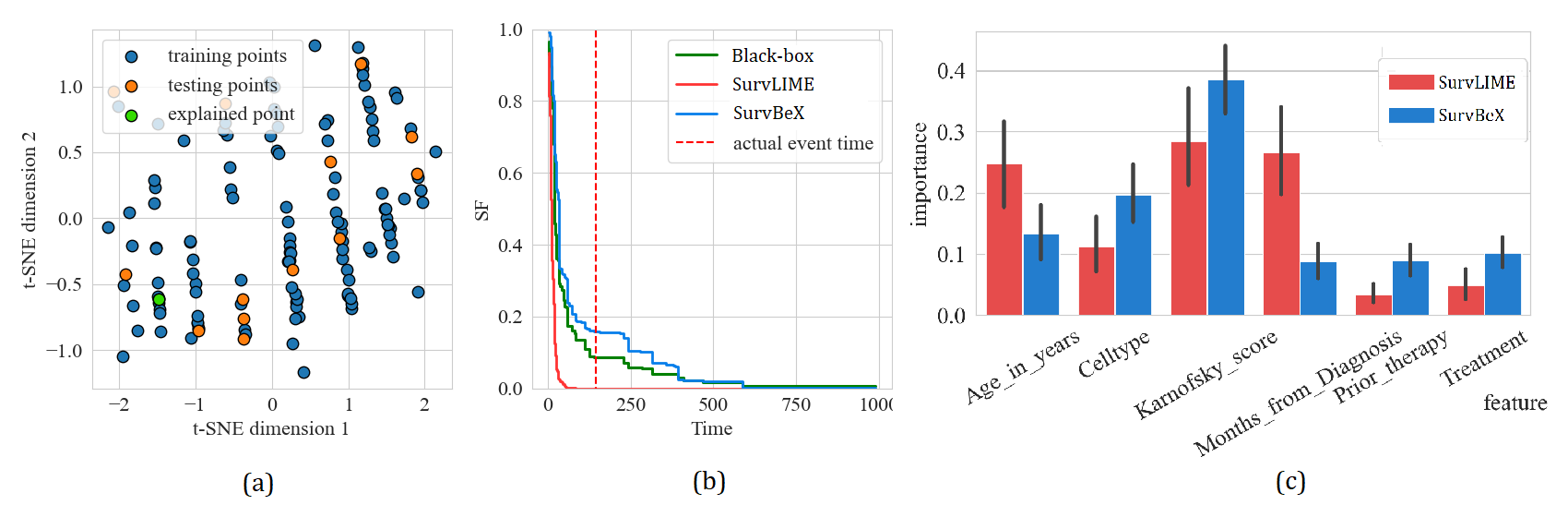}%
\caption{Experiments with the Veteran dataset: (a) the t-SNE representation of
data; (b) SFs predicted by the black-box model, SurvLIME and SurvBeX; (c) mean
values of the feature importance}%
\label{f:real_veteran}%
\end{center}
\end{figure}

Fig.\ref{f:real_veteran}(a) shows the data structure of the dataset Veteran by
means of the t-SNE method. Fig.\ref{f:real_veteran}(b) shows SFs predicted by
the black-box model, SurvLIME and SurvBeX. It can be seen from
Fig.\ref{f:real_veteran} that the dataset has a specific structure which does
not allows us to construct a proper Cox model in the framework of SurvLIME. As
a result, the SF produced by SurvBeX is closer to the SF predicted by the
black-box model than the SF produced by SurvLIME. Mean values of the feature
importance over all points of the dataset computed by using SurvLIME\ and
SurvBeX are depicted in Fig.\ref{f:real_veteran}(c). It can be seen from
Fig.\ref{f:real_veteran}(c) that the results provided by SurvLIME\ and SurvBeX
are different. The methods coincide in selection of the most important feature
\textquotedblleft Karnofsky score\textquotedblright. However, the most
important second feature is determined by SurvBeX as \textquotedblleft
Celltype\textquotedblright \ whereas SurvLIME determines it as
\textquotedblleft Months from diagnosis\textquotedblright. \textquotedblleft
Karnofsky score\textquotedblright \ and \textquotedblleft
Celltype\textquotedblright \ as the most important features were determined
also in \cite{Utkin-Satyukov-Konstantinov-22}.%

\begin{figure}
[ptb]
\begin{center}
\includegraphics[
height=1.9331in,
width=5.9329in
]%
{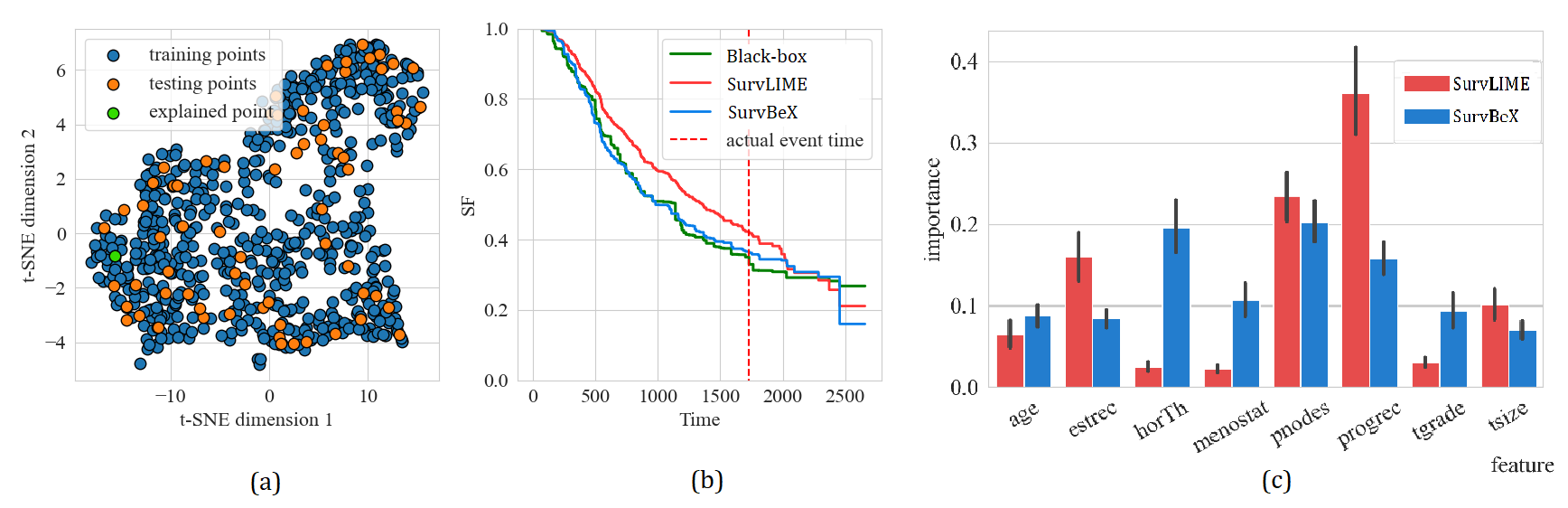}%
\caption{Experiments with the GBSG2 dataset: (a) the t-SNE representation of
data; (b) SFs predicted by the black-box model, SurvLIME and SurvBeX; (c) mean
values of the feature importance}%
\label{f:real_gbsg}%
\end{center}
\end{figure}

The next dataset for studying is GBSG2. The data structure of the dataset
GBSG2 depicted by means of the t-SNE method and SFs predicted by the black-box
model SurvLIME and SurvBeX are shown in Fig.\ref{f:real_gbsg}(a) and (b),
respectively, We again see that the SF produced by SurvBeX is closer to the SF
predicted by the black-box model than the SF produced by SurvLIME.
Fig.\ref{f:real_gbsg}(c) shows mean values of the feature importance over all
points of the dataset GBSG2 computed by using SurvLIME\ and SurvBeX.

\begin{figure}
[ptb]
\begin{center}
\includegraphics[
height=2.0316in,
width=5.8792in
]%
{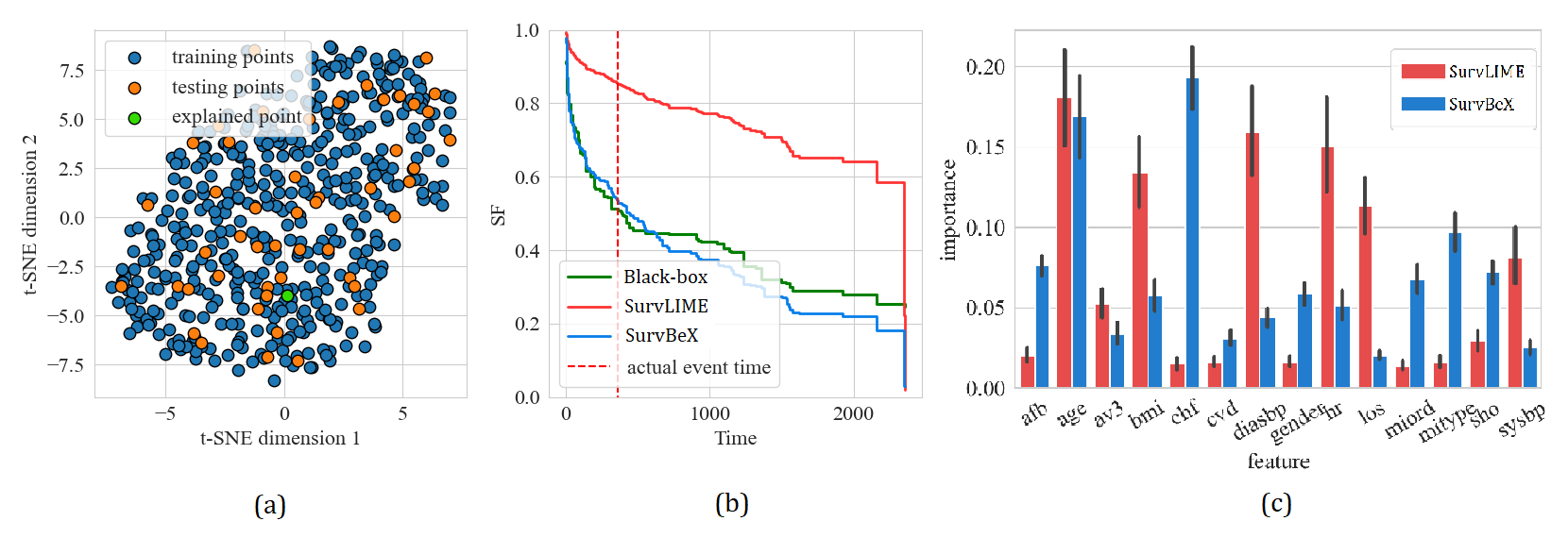}%
\caption{Experiments with the WHAS500 dataset: (a) the t-SNE representation of
data; (b) SFs predicted by the black-box model, SurvLIME and SurvBeX; (c) mean
values of the feature importance}%
\label{f:real_whas500}%
\end{center}
\end{figure}

The same results can be seen in Fig.\ref{f:real_whas500} where results
obtained for the dataset WHAS500 are shown.%

%

\section{Conclusion}

A new explanation method called SurvBeX, which can be regarded as an
alternative to the well-known method SurvLIME developed for survival data, has
been presented in the paper. The main idea behind the method is to use the
Beran estimator to approximate a survival black-box model at an explained
point. The explanation by means of SurvBeX is reduced to minimization between
SFs predicted by the black-box model and the Beran estimator for points
generated around the explained point.

There is an important difference between SurvLIME and SurvBeX except for the
use of the Beran estimator instead of the Cox model. According to SurvLIME,
the importance weights $\mathbf{b}$ are assigned to features. In SurvBeX, they
are assigned to distances between features of the explained example and
training examples. This implies that the importance weights in SurvBeX
directly depend on the training set whereas the weights in SurvLIME implicitly
depend on the training set through the baseline function which may be
incorrect when the analyzed dataset has a complex structure.

At the same time, we should point out a drawback of SurvBeX. In contrast to
SurvLIME, SurvBeX has to solve the complex optimization problem, which leads
to increasing the computation time. Nevertheless, various numerical
experiments have demonstrated that the optimization problem is successfully
solved and SurvBeX provides the outperforming explanation.

There are several ways for further research based on the proposed method.
First of all, we have considered the minimization of the distances between
SFs. In the same way, the cumulative hazard functions can be considered as it
has been implemented in SurvLIME. Second, only the Gaussian kernel and its
modification have been used to implement SurvBeX. However, it is interesting
to consider other types of kernels which may provide outperforming results.
Third, another important generalization of the proposed method is to combine
SurvLIME and SurvBeX. For example, they can be linearly combined with some
parameter such that the SF of the combined explanation model is represented as
a linear combination of two SFs with the same parameters $\mathbf{b}$. The
above problems are directions for further research.

\section*{Acknowledgement}

The research is partially funded by the Ministry of Science and Higher
Education of the Russian Federation as part of state assignments
\textquotedblleft Development of a Multi-Agent Resource Manager for a
Heterogeneous Supercomputer Platform Using Machine Learning and Artificial
Intelligence\textquotedblright \ (topic FSEG-2022-0001).

\bibliographystyle{unsrt}
\bibliography{Classif_bib,Deep_Forest,Explain,Explain_med,IntervalClass,MYBIB,MYUSE,Survival_analysis}

\end{document}